\documentclass[10pt,twocolumn,letterpaper]{article}

\usepackage[pagenumbers]{cvpr} % To force page numbers, e.g. for an arXiv version

\definecolor{cvprblue}{rgb}{0.21,0.49,0.74}
\usepackage[pagebackref,breaklinks,colorlinks,allcolors=cvprblue]{hyperref}

\def\paperID{1336} % *** Enter the Paper ID here
\def\confName{CVPR}
\def\confYear{2025}

\usepackage[utf8]{inputenc} % allow utf-8 input
\usepackage{CJKutf8} % allow chinese input
\usepackage{mathrsfs}
\usepackage{amsmath}
\usepackage{bm}
\usepackage{multirow}
\usepackage{bbding}
\usepackage{tcolorbox}
\usepackage[dvipsnames]{xcolor}
\hypersetup{
	colorlinks,
	urlcolor={purple}
}

\definecolor{rred}{RGB}{245, 152, 153}
\definecolor{oorange}{RGB}{253, 205, 154}

\title{MatAnyone: Stable Video Matting with Consistent Memory Propagation}

\author{Peiqing Yang$^{1}$ \quad Shangchen Zhou$^{1}$ \quad Jixin Zhao$^{1}$ \quad Qingyi Tao$^{2}$ \quad Chen Change Loy$^{1}$\\
$^{1}$S-Lab, Nanyang Technological University \quad $^{2}$SenseTime Research, Singapore\\
{\tt\small \url{https://pq-yang.github.io/projects/MatAnyone}}
}

\begin{document}
\twocolumn[{%
\renewcommand\twocolumn[1][]{#1}%
\maketitle
\vspace{-9mm}
\begin{center}
    \centering
    \includegraphics[width=\linewidth]{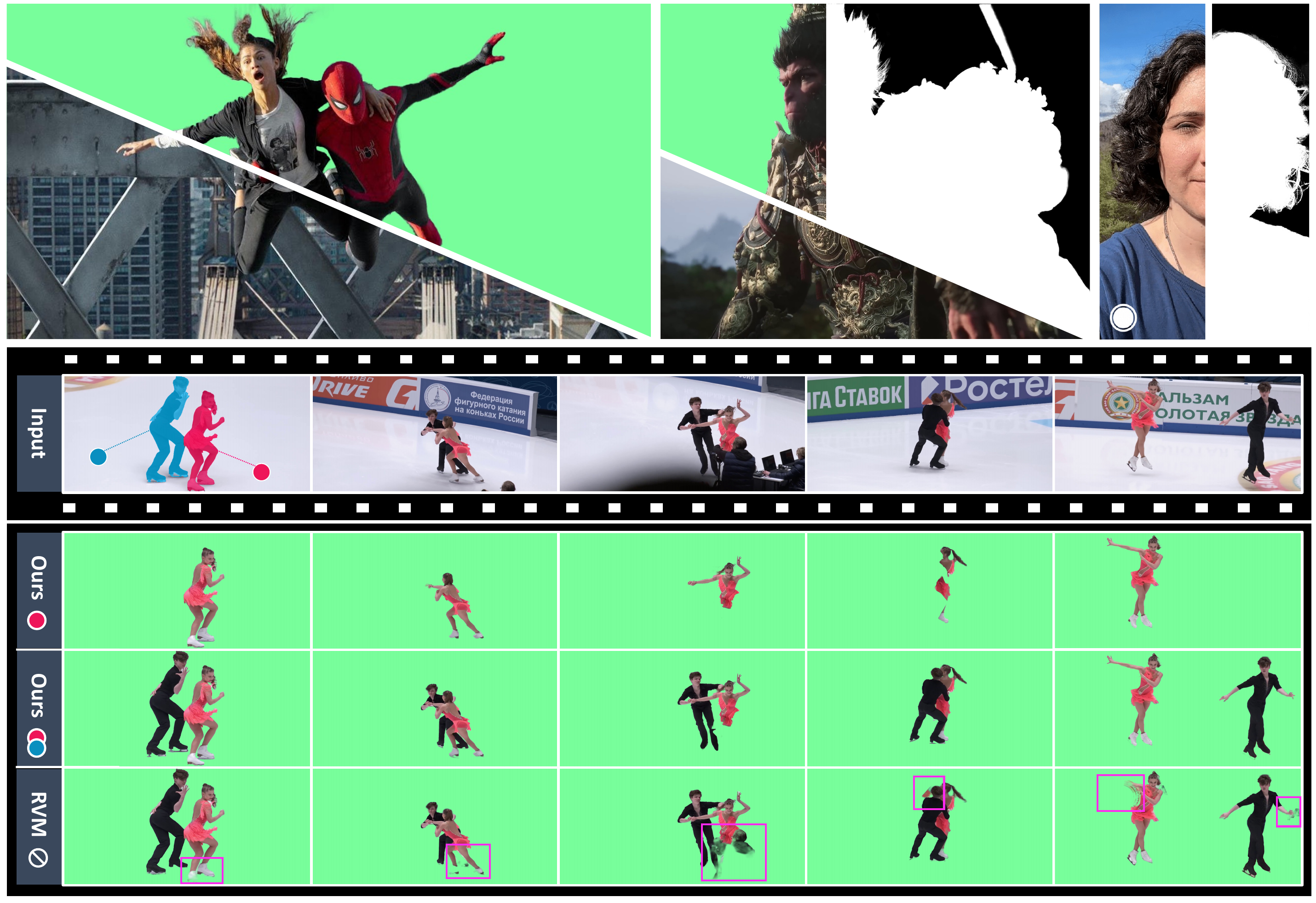}
    \vspace{-6mm}
    \captionof{figure}{%
    % Our \textit{MatAnyone} demonstrates its ability in both presenting high-quality detail extraction (upper part) and consistent semantic tracking ability (lower part).
    %
    % (a) Our method is able to solve diverse frame sizes (\eg, wide, medium, and close-up) and diverse media types (\eg, movies, games, cellphone videos), presenting details at the image-matting level.
    %
    % (b) While robust auxiliary-free methods like RVM~\cite{} struggle with complex backgrounds (e.g., similar colors or multiple humans in the scene), our approach effectively differentiates the target human from distractors by maintaining a clean background and complete body parts. Furthermore, our method excels at consistently tracking the target object even in the presence of multiple salient objects, accurately distinguishing body parts during interactions.
    %
    Our \textit{MatAnyone} is capable of producing highly detailed and temporally consistent alpha mattes throughout a video.    
    % achieves high-quality detail extraction (top) and consistent semantic tracking of the target object pre-assigned in the first frame (bottom).
    %
    % (a) It adapts to a variety of frame sizes (\eg, wide, medium, close-up) and media types (\eg, films, games, smartphone videos), achieving fine-grained details at the image-matting level.
    (a) It adapts to a variety of frame sizes and media types (\eg, films, games, smartphone videos), achieving fine-grained details at the image-matting level.
    %
    % (b) RVM~\cite{lin2022rvm}, an auxiliary-free video matting method struggles with complex (\eg, confusing patterns) or ambiguous (\eg, multiple people) backgrounds. In contrast, our method effectively isolates the target object from such distractors, preserving a clean background and complete foreground body parts.
    (b) RVM~\cite{lin2022rvm}, an auxiliary-free video matting method, struggles with complex or ambiguous backgrounds. In contrast, our method effectively isolates the target object from such distractors, preserving a clean background and complete foreground parts.
    %
    % (c) Our method also excels at tracking the target (\ie, the lady in pink) consistently, even in scenes with multiple objects (\ie, the man and the lady), accurately distinguishing their body parts during interactions.
    (c) Our method also excels at consistently tracking the target (\ie, the lady in pink) even in scenes containing multiple salient objects (\ie, the man and the lady). It accurately distinguishes between them even during their interactions.
    \textbf{(Zoom-in for best view)}
    } \vspace{1mm}
    \label{fig:teaser}
\end{center}%
}]

\begin{abstract}
Auxiliary-free human video matting methods, which rely solely on input frames, often struggle with complex or ambiguous backgrounds. To address this, we propose MatAnyone, a robust framework tailored for target-assigned video matting. Specifically, building on a memory-based paradigm, we introduce a consistent memory propagation module via region-adaptive memory fusion, which adaptively integrates memory from the previous frame. This ensures semantic stability in core regions while preserving fine-grained details along object boundaries. For robust training, we present a larger, high-quality, and diverse dataset for video matting. Additionally, we incorporate a novel training strategy that efficiently leverages large-scale segmentation data, boosting matting stability. With this new network design, dataset, and training strategy, MatAnyone delivers robust and accurate video matting results in diverse real-world scenarios, outperforming existing methods. 
% The code and model will be publicly available.
\end{abstract}    
\section{Introduction}
\label{sec:intro}

\begin{figure*}[t]
\begin{center}
    \vspace{-5mm}
    \includegraphics[width=\linewidth]{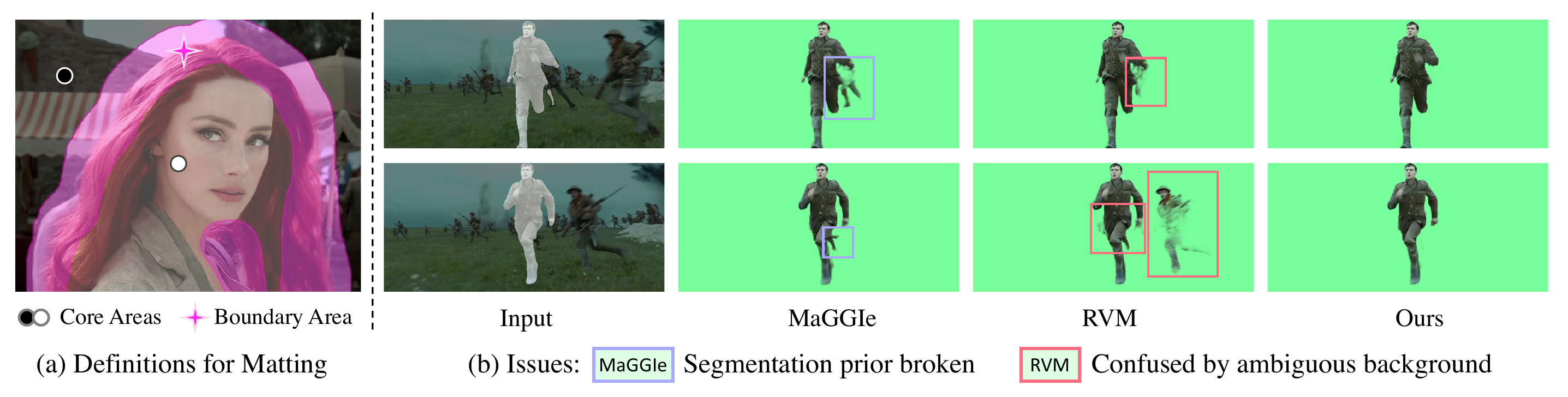}
    \vspace{-7mm}
    \caption{
    Definitions and motivations for MatAnyone.
    (a) In a matting frame, the image can be broadly divided into two areas based on the alpha value: the \textit{core} (semantic) and the \textit{boundary} (fine-details). The \textit{core} includes the background (alpha values of 0) and the solid foreground (alpha values of 1), while the \textit{boundary} (highlighted in pink) encompasses areas with alpha values between 0 and 1.
    (b) Due to the under-defined setting, auxiliary-free methods like RVM~\cite{lin2022rvm} are easily confused by ambiguous background. Meanwhile, mask-guided methods like MaGGIe~\cite{huynh2024maggie} tend to break the segmentation prior they aim to leverage, due to the deficiency in video matting data.
    }
    \vspace{-7mm}
\label{fig:motivation}
\end{center}
\end{figure*}

% why this setting
%
Auxiliary-free human video matting (VM) is widely recognized for its convenience~\cite{lin2022rvm,ke2022MODNet,li2024vmformer}, as it only requires input frames without additional annotations. However, its performance often deteriorates in complex or ambiguous backgrounds, especially when similar objects, \ie, other humans, appear in the background (Fig.~\ref{fig:motivation}(b)). We consider auxiliary-free video matting to be under-defined, as their results can be uncertain without a clear target object.

In this work, we focus on a problem that is more applicable to real-world video applications: video matting focused on pre-assigned target object(s), with the target segmentation mask provided in the first frame. This enables the model to perform stable matting via consistent object tracking throughout the entire video, while offering better interactivity.
%
% why memory-based
%
The setting is well-studied in Video Object Segmentation (VOS), where it is referred to as ``semi-supervised"~\cite{hu2017maskrnn, oh2019video, cheng2021stcn}. A common strategy is to use a memory-based paradigm~\cite{oh2019video, xie2021efficient, cheng2022xmem, cheng2024cuite}, encoding past frames and corresponding segmentation results into memory, from which a new frame retrieves relevant information for its mask prediction. This allows a lightweight network to achieve consistent and accurate tracking of the target object. Inspired by this, we adapt the memory-based paradigm for video matting, leveraging its stability across frames.
%

% current practice
%
Video matting poses additional challenges compared to VOS, as it requires not only \textit{accurate semantic detection} in \textit{core} regions but also \textit{high-quality detail extraction} along the \textit{boundary} (\eg, hair), as defined in Fig.~\ref{fig:motivation}(a). A straightforward approach is to fine-tune matting details using matting data, based on segmentation priors from VOS. Recent approaches attempt to achieve both goals, either in a coupled or decoupled manner. For instance, AdaM~\cite{lin2023adam} and FTP-VM~\cite{huang2023ftp} refine the memory-based segmentation mask for each frame via a decoder to produce alpha mattes, while MaGGIe~\cite{huynh2024maggie} devises a separate refiner network to process segmentation masks across all frames from an off-the-shelf VOS model. However, these methods often lead to suboptimal results due to limitations in the available video matting data: 
(i)~the quality of VideoMatte240K~\cite{lin2021bgm}, the most widely used human video matting dataset, is suboptimal. Its ground-truth alpha mattes exhibit problematic semantic accuracy in core areas (\eg, interior holes) and lack fine details along the boundaries (\eg, blurry hair);
(ii)~video matting datasets are much smaller in scale compared to VOS datasets; 
and (iii)~video matting data are synthetic due to the extreme difficulty of human annotations, limiting their generalizability to real-world cases~\cite{lin2022rvm}.
Consequently, fine-tuning a strong VOS prior for video matting with existing video matting data usually disrupts this prior. While boundary details may show improvement compared to segmentation results, the matting quality in terms of semantic stability in \textit{core} areas and details in \textit{boundary} areas remain unsatisfactory, as shown by the results of MaGGIe in Fig.~\ref{fig:motivation}(b).
%

% our designs
%
Producing matting-level details while maintaining semantic stability of a memory-based approach is challenging, especially training with suboptimal video matting data. To tackle this, we focus on several key aspects:

\noindent\textbf{Network} - we introduce a \textit{consistent} memory propagation mechanism in the memory space.
For each current frame, the alpha value change relative to the previous frame is estimated for every token. This estimation guides the adaptive integration of information from the previous frame.
The ``large-change'' regions rely more on the current frame’s information queried from the memory bank, while ``small-change'' regions tend to retain the memory from the previous frame. This region-adaptive memory fusion inherently stabilizes memory propagation throughout the video, improving matting quality with fine details and temporal consistency. Specifically, it encourages the network to focus on boundary regions during training to capture fine details, while ``small-change'' tokens in the core regions preserve internally complete foreground and clean background (see our results in Fig.~\ref{fig:motivation}(b)).
\noindent\textbf{Data} - we collect a new training dataset, named \textit{VM800}, which is twice as large, more diverse, and of higher quality in both core and boundary regions compared to the VideoMatte240K dataset~\cite{lin2021bgm}, greatly enhancing robust training for video matting. In addition, we introduce a more challenging test dataset, named \textit{YoutubeMatte}, featuring more diverse foreground videos and improved detail quality. These new datasets offer a solid foundation for robust training and reliable evaluation in video matting.
\noindent\textbf{Training Strategy} - the lack of \textit{real} video matting data remains a significant limitation, affecting both stability and generalizability. We address this problem by leveraging large-scale \textit{real} segmentation data via a novel training strategy. Unlike common practices~\cite{lin2022rvm,huynh2024maggie,huang2023ftp} that train with segmentation data on a separate prediction head parallel to the matting head, we propose using segmentation data within the same head as matting for more effective supervision. This is achieved by applying region-specific losses -- for \textit{core} regions, we apply a pixel-wise loss to ensure stability and generalization in semantics; for \textit{boundary} regions, where segmentation data lacks alpha labels, we employ an improved DDC loss~\cite{liu2024ddc}, scaled to make edges resemble matting rather than segmentation.

% summary
%
In summary, our main contributions are as follows:
\begin{itemize}
    \item We propose \textit{MatAnyone}, a practical human video matting framework supporting target assignment, with stable performance in both semantics of \textit{core} regions and fine-grained \textit{boundary} details. Target object(s) can be easily assigned using off-the-shelf segmentation methods, and reliable tracking is achieved even in long videos with complex and ambiguous backgrounds.
    \item We introduce a \textit{consistent} memory propagation mechanism via \textit{region-adaptive} memory fusion, improving stability in \textit{core} regions and quality in \textit{boundary} details.
    \item We contribute larger and higher-quality datasets for training and testing, offering a solid foundation for robust training and reliable evaluation in video matting.
    \item To overcome the scarcity of real video matting data, we leverage real segmentation data for core-area supervision, largely improving semantic stability over prior methods.
\end{itemize}

\section{Related Work}
\label{sec:relatedwork}

\noindent \textbf{Video Matting.}
Due to the intrinsic ambiguity in the auxiliary-free setting~\cite{zhang2019late,qiao2020attention,lin2022rvm,ke2022MODNet,li2024vmformer,zhu2017fast}, such tasks generally are object-specific. Among them, human video matting~\cite{shen2016deep,zhu2017fast,ke2022MODNet,li2024vmformer} without auxiliary inputs is popular due to its wide applications. 
Challenging as the auxiliary-free setting, being in the video domain brings in additional difficulties in temporal coherency.
MODNet~\cite{ke2022MODNet} extends its portrait matting setting to video domain with a flickering reduction trick (non-learning) within a local sequence. RVM~\cite{lin2022rvm} steps further to design for videos specifically with ConvGRU~\cite{ballas2016convgru} as its recurrent architecture. 
Robust as RVM, it is still easy to be confused by humans in the background.
With the success of promptable segmentation~\cite{kirillov2023sam,ravi2024sam2,zou2024seem,zhou2023edgesam}, obtaining segmentation mask for a target human object only requires minimal human efforts. Recent mask-guided image~\cite{yao2024matteanything,yao2024vitmatte,cai2022transmatting,li2024matting} and video matting~\cite{huang2023ftp, lin2023adam, huynh2024maggie,li2024vim} thus leverage this convenience for a more robust performance.
Adam~\cite{lin2023adam} propagates the first-frame segmentation mask across all frames while FTP-VM~\cite{huang2023ftp} propagates the first-frame trimap. Taking the propagated mask as a rough result, their decoder serves for matting details refinement. MaGGIe~\cite{huynh2024maggie} enjoys a stronger prior by taking the segmentation mask across all frames instead of the first one. Taking all the segmentation masks at a time, the network is able to perform bidirectional temporal fusion for coherency. 
To mitigate the poor generalizability of synthetic video matting data, a common practice is to simultaneously train with real segmentation data for semantic supervision~\cite{lin2022rvm,lin2023adam,huang2023ftp}.

\noindent \textbf{Memory-based VOS.}
Semi-supervised VOS segments the target object with a first-frame annotation across frames~\cite{oh2019stm,cheng2021stcn,cheng2022xmem, cheng2024cuite, cheng2023tracking, li2023tube, seong2020kernelized, cheng2021modular, hu2021learning}.
The memory matching paradigm by Space-Time Correspondence Network (STCN)~\cite{cheng2021stcn} is widely followed by current VOS methods~\cite{cheng2022xmem,cheng2024cuite,xie2021efficient,wang2021swiftnet}, and achieves good performance. We thus take the memory-based paradigm as our framework since it is similar to our setting except that our outputs are alpha mattes.

\noindent \textbf{Video Consistency in Low-level Vision.}
To enhance temporal consistency across adjacent frames, the recurrent frame fusion~\cite{zhou2019stfan, wang2019edvr} and optical flow-guided propagation~\cite{chan2021basicvsr, chan2022basicvsr++, chan2022realbasicvsr, zhou2023propainter} are commonly utilized in the video restoration networks.
Recent methods also employ temporal layers such as 3D convolution~\cite{blattmann2023align, wang2023videocomposer} and temporal attention~\cite{blattmann2023align, chen2023videocrafter1, zhou2023upscale, wang2024lavie} during training, while other training-free methods resort to cross-frame attention~\cite{wu2023tune, yang2023rerender} and flow-guided attention~\cite{geyer2023tokenflow, cong2023flatten} in the pretrained models.
In this work, we find that the memory-based paradigm is effective enough to maintain video consistency for video matting.
\section{Methodology}
\label{sec:method}

\begin{figure*}[t]
\begin{center}
    \vspace{-8mm}
    \includegraphics[width=.98\linewidth]{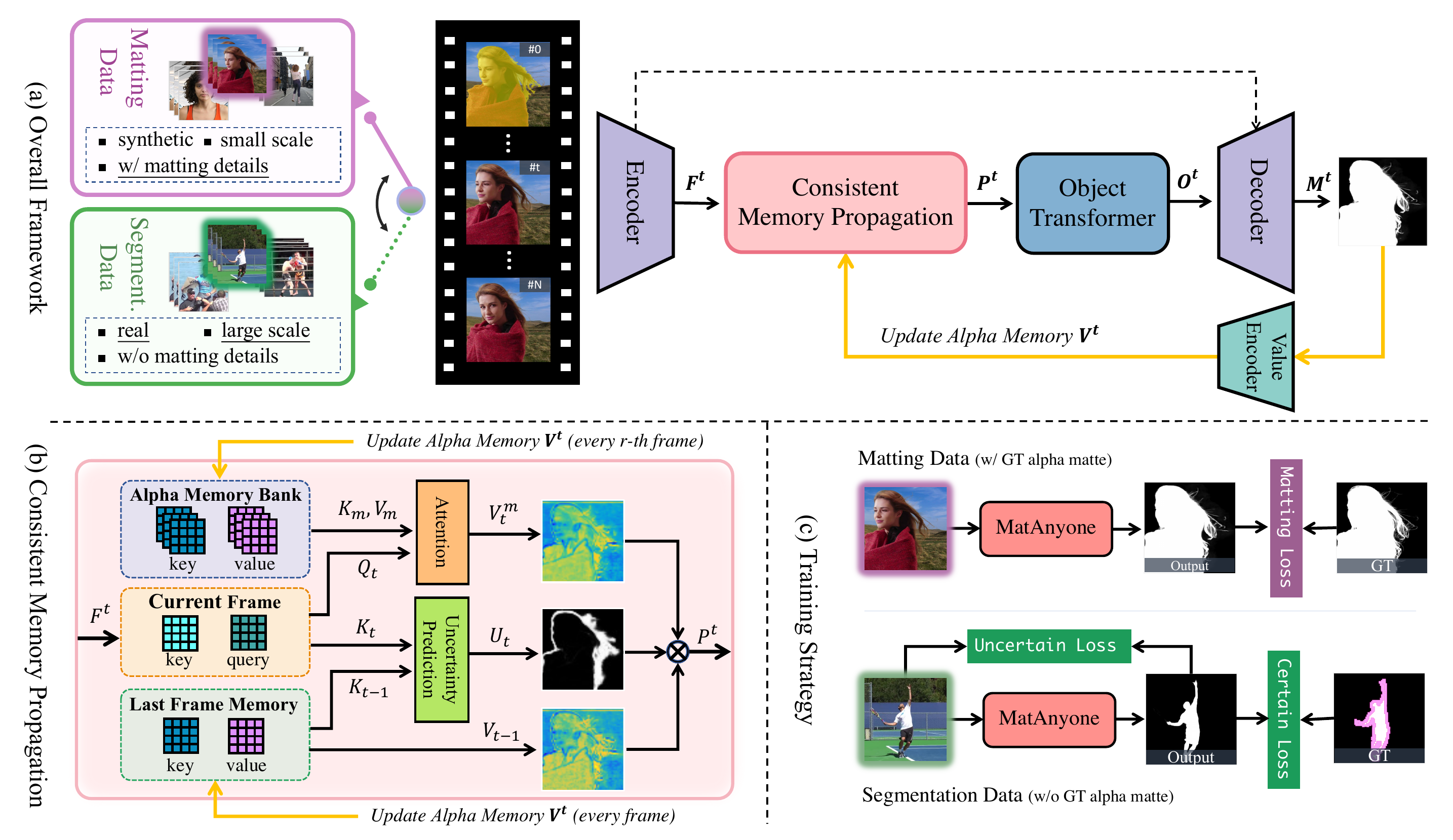}
    \vspace{-3mm}
    \caption{
    An overview of MatAnyone.
    MatAnyone is a memory-based framework for video matting. Given a target segmentation map in the first frame, our model achieves stable and high-quality matting through consistent memory propagation, with a region-adaptive memory fusion module to combine information from the previous and current frame. To overcome the scarcity of real video matting data, we incorporate a new training strategy that effectively leverages matting data for fine-grained matting details and segmentation data for semantic stability, with designed losses separately.
    }
    \vspace{-6mm}
\label{fig:overview}
\end{center}
\end{figure*}

\noindent \textbf{Overview.} Achieving matting-level details while preserving the semantic stability of a memory-based approach poses challenges, especially when training with suboptimal video matting data. To tackle this, we propose our \textit{MatAnyone}, as illustrated in Fig.~\ref{fig:overview}.
Similar to semi-supervised VOS, MatAnyone only requires the segmentation mask for the first frame as a target assignment (\eg, the yellow mask in Fig.~\ref{fig:overview}(a)). The alpha matte for the assigned object is then generated frame by frame in a sequential manner.
Specifically, for an incoming frame $t$, it is first encoded into $F^t$ as $\times 16$ downsampled feature representation, which is then transformed into key and query for consistent memory propagation (Sec.~\ref{subsec:mem_prop}), and output the pixel memory readout $P^t$. We employ the object transformer proposed by Cutie~\cite{cheng2024cuite} to group the pixel memory by object-level semantics for robustness against noise brought by low-level pixel matching. The refined memory readout $O^t$ acts as the final feature to be sent into the decoder for alpha matte prediction. The predicted alpha matte $M^t$ is then encoded to memory value $V^t$, which is used to update the alpha memory bank.
Due to limitations in the quality and quantity of video matting data, training with such data makes it difficult to achieve satisfactory stability in core regions. To mitigate this, RVM~\cite{lin2022rvm} proposes a parallel head for \textit{real} segmentation data alongside the matting head, guiding the network to be robust in real-world cases.
However, this is not sufficient, as the matting head itself cannot receive supervision from real data. Inspired by the DDC loss~\cite{liu2024ddc} designed for alpha-free image matting, we devise a training strategy for core regions, which provides \textit{direct} supervision to the matting head with segmentation data (Sec.~\ref{subsec:cert_sup}), leading to substantial improvements in semantic stability.
We also propose a practical inference strategy that allow for flexible application: a recurrent refinement approach applied to the first frame, based on the memory-driven paradigm, enhancing robustness to the given mask and refining matting details (Sec.~\ref{subsec:infer_strategy}).

\subsection{Consistent Memory Propagation}
\label{subsec:mem_prop}
\noindent \textbf{Alpha Memory Bank.} 
In this study, we introduce a \textit{consistent memory propagation} (CMP) module specifically designed for video matting, as illustrated in Fig.\ref{fig:overview}(b). Existing memory-based VM methods store either segmentation masks~\cite{lin2023adam} or trimaps~\cite{huang2023ftp} in memory and use a decoder to refine the matting details. Such approaches do not fully leverage the stability provided by the memory paradigm in boundary regions, leading to instability such as flickering. To address this, building on the memory-based framework~\cite{cheng2021stcn}, our MatAnyone stores the alpha matte in an alpha memory bank to enhance stability in boundary regions.
\noindent \textbf{Region-Adaptive Memory Fusion.} %
Given the inherent difference between the segmentation map (values of 0 or 1) and the matting map (values between 0 and 1), the memory-matching approach needs to be adjusted. 
Specifically, in STCN~\cite{cheng2021stcn}, memory values for the query frame are based on the similarity between query and memory key, assuming equal importance for all query tokens. However, this assumption does not hold for video matting. As shown in Fig.~\ref{fig:motivation}(a), a query frame can be divided into \textit{core} and \textit{boundary} regions. When compared with frame $t-1$, only a small fraction of tokens in frame $t$ change significantly in alpha values, with these ``large-change" tokens mainly located in object boundaries, while the ``small-change" tokens reside in the core regions. This highlights the need to treat \textit{core} and \textit{boundary} regions separately to enforce stability.
Specifically, we introduce a boundary-area prediction module to estimate the change probability $U_t$ of each query token for adaptive memory fusion, where higher $U_t$ indicates ``large-change" regions and lower $U_t$ indicates ``small-change" regions.
The prediction module is lightweight, consisting of three convolution layers. We formulate the prediction as a binary segmentation problem with loss $\mathcal{L}_{{bin\_seg}}$ and use the actual alpha change between frame ${t-1}$ and $t$ as supervision. 
Specifically, we define $U^{GT}_t: |M^{GT}_{t-1} - M^{GT}_{t}| >= \delta $, where $\delta$ is a threshold.
Using the output of the module $\hat{U}_t$, we compute the binary cross entropy loss against $U^{GT}_t$. During the region-adaptive memory fusion process, we apply the sigmoid function on $\hat{U}_t$ to transform it as a probability.
The final pixel memory readout is a soft merge:
\begin{equation}
P_t = V^m_{t} * U_t + V_{t-1} * (1 - U_t),
\label{eq:merge_value}
\end{equation}
where $U_t \in [0, 1]$, $V^m_{t}$ are current values queried from memory bank, and $V_{t-1}$ are values propagated from the last frame. 
This approach significantly improves stability in core regions by maintaining internal completeness and a clean background (Fig.~\ref{fig:motivation}(b) and Fig.~\ref{fig:qualitative_real}). It also enhances stability in boundary regions, as it directs the network to focus on object boundaries with soft alpha values, while the memory-based paradigm inherently stabilizes the matched values (see Table~\ref{tab:ablation}(c)). A detailed analysis is provided in the ablation study of Sec.~\ref{sec:ablation} and Sec.~\ref{suppl_subsec:cmp}.
\subsection{Core-area Supervision via Segmentation}
\label{subsec:cert_sup}

\noindent \textbf{New Training Scheme.}
Most recent video matting methods follow RVM's approach of using real segmentation data to address the limitations of video matting data. In these methods, segmentation and matting data are fed to the main shared network, but are directed to produce outputs at separate heads. 
Although segmentation data do supervise the main network to empower generalizability and robustness to the matting model, the stability they provide falls short of what a VOS model could achieve. As shown in Fig.~\ref{fig:motivation}, both RVM and MaGGIe perform significantly worse than the VOS outputs (white masks on inputs) by XMem~\cite{cheng2022xmem} in core areas, where semantic information is key.
We believe the parallel head training scheme may not fully exploit the rich segmentation prior in the data. To address this, we propose to supervise the matting head directly with segmentation data. Specifically, we predict the \textit{alpha matte} for segmentation inputs and optimize the matting outputs accordingly, as illustrated in Fig.~\ref{fig:overview}(c).
\noindent \textbf{Scaled DDC Loss.} 
A natural challenge arises with the aforementioned approach: how can we compute the loss on matting outputs for segmentation data when there is no ground truth (GT) alpha matte? For core areas, the GT labels are readily available in the segmentation data, where an $l1$ loss suffices, and we denote it as $\mathcal{L}_{core}$.
The real difficulty lies in the boundary region. A recent paper proposes DDC loss~\cite{liu2024ddc}, which supervises boundary areas using the input image without requiring a GT alpha matte.
\begin{equation}
\begin{array}{lr}
\label{eq:ddc}
    &\mathcal{L}_{DDC} = \frac{1}{N} \sum\limits^N_i \sum\limits_j |\alpha_i - \alpha_j - \| \bm{I}_i - \bm{I}_j \|_2|, \\
    &j \in \text{argtopk}\{-\| \bm{I}_i - \bm{I}_j \|_2\}.
\end{array}
\end{equation}
However, we find that the underlying assumption of this design, that $\| \alpha_i - \alpha_j \|_2 = \| \bm{I}_i - \bm{I}_j \|_2$ for $\alpha_i > \alpha_j$, does not always hold true.
For two image pixels $\bm{I}_i$ and $\bm{I}_j$, their difference is given by:
\begin{equation}
\label{eq:derive}
    \bm{I}_i - \bm{I}_j = [\alpha_iF_i + (1-\alpha_i)B_i] - [\alpha_jF_j + (1-\alpha_j)B_j],
\end{equation}
where $F_i$, $B_i$ represent the foreground and background values at pixel $i$, and similarly for $F_j$ and $B_j$ at pixel $j$. Since we impose the constraint  $j \in \text{argtopk}\{-\| \bm{I}_i - \bm{I}_j \|_2\}$, we can assume $F_i=F_j=F$, $B_i=B_j=B$ within a small window. This simplifies Eq.~(\ref{eq:derive}) to:
\begin{equation}
    \bm{I}_i - \bm{I}_j = (\alpha_i-\alpha_j)(F-B).
\end{equation}
This shows that the assumptions for DDC loss hold only when $|F-B|=1$. To account for this, we devise a scaled version as our boundary loss $\bm{L}_{boundary}$:
\begin{equation}
\begin{array}{r}
\label{eq:scaled_ddc}
    \mathcal{L}_{boundary} = \frac{1}{N} \sum\limits^N_i \sum\limits_j |(\alpha_i - \alpha_j)\mathbf{(F-B)} - \| \bm{I}_i - \bm{I}_j \|_2|, \\
    j \in \text{argtopk}\{-\| \bm{I}_i - \bm{I}_j \|_2\},
\end{array}
\end{equation}
where $F$ is approximated by the average of the top $k$ largest pixel values in the small window, and $B$ by the average of the top $k$ smallest pixel values.
In the ablation study (Sec.~\ref{sec:ablation}), we show that training with our scaled DDC loss (Eq.~(\ref{eq:scaled_ddc})) yields more natural matting results than training with the original version (Eq.~(\ref{eq:ddc})), which tends to produce segmentation-like jagged and stair-stepped edges.

\begin{table*}[t!]
\begin{center}
\vspace{-6mm}
\setlength{\fboxsep}{2.8pt}
\caption{
    Quantitative comparisons on different video matting benchmarks from diverse sources. The best and second-best performances are marked in \colorbox{rred}{\underline{red}} and \colorbox{oorange}{orange}, respectively.
    $\dag$ indicates that MaGGIe~\cite{huynh2024maggie} requires the instance mask as guidance for each frame, while our method only requires it in the first frame.
    }
\label{tab:comparison_syn}
\vspace{-1mm}
\renewcommand{\arraystretch}{1.2}
\renewcommand{\tabcolsep}{4.5mm}
\scalebox{0.84} {
\begin{tabular}{lccccccc}
\hline
\multicolumn{1}{l|}{\multirow{2}{*}{Metrics}} &
  \multicolumn{3}{c|}{\textbf{Auxiliary-free (AF) Methods}} &
  \multicolumn{4}{c}{\textbf{Mask-guided Methods}} \\ \cline{2-8} 
\multicolumn{1}{l|}{} &
  MODNet~\cite{ke2022MODNet} &
  RVM~\cite{lin2022rvm} &
  \multicolumn{1}{c|}{RVM-Large~\cite{lin2022rvm}} &
  AdaM~\cite{lin2023adam} &
  FTP-VM~\cite{huang2023cvpr} &
  MaGGIe$^{\dag}$~\cite{huynh2024maggie} &
  \textbf{Ours} \\ \hline \hline
\multicolumn{8}{l}{\textbf{\textit{VideoMatte}} ($512 \times 288$)} \\ \hline
\multicolumn{1}{l|}{MAD$\downarrow$} &
  \multicolumn{1}{c}{9.41} &
  \multicolumn{1}{c}{6.08} &
  \multicolumn{1}{c|}{5.32} &
  \multicolumn{1}{c}{\colorbox{oorange}{{5.30}}} &
  \multicolumn{1}{c}{6.13} &
  \multicolumn{1}{c}{5.49} &
  \multicolumn{1}{c}{\colorbox{rred}{\underline{5.15}}} \\
\multicolumn{1}{l|}{MSE$\downarrow$} &
  4.30 &
  1.47 &
  \multicolumn{1}{c|}{\colorbox{oorange}{0.62}} &
  0.78 &
  1.31 &
  \colorbox{rred}{\underline{0.60}} &
  0.93 \\
\multicolumn{1}{c|}{Grad$\downarrow$} &
  1.89 &
  0.88 &
  \multicolumn{1}{c|}{\colorbox{oorange}{0.59}} &
  0.72 &
  1.14 &
  \colorbox{rred}{\underline{0.57}} &
  0.67 \\
\multicolumn{1}{l|}{dtSSD$\downarrow$} &
  \multicolumn{1}{c}{2.23} &
  \multicolumn{1}{c}{1.36} &
  \multicolumn{1}{c|}{\colorbox{oorange}{1.24}} &
  \multicolumn{1}{c}{1.33} &
  \multicolumn{1}{c}{1.60} &
  \multicolumn{1}{c}{1.39} &
  \multicolumn{1}{c}{\colorbox{rred}{\underline{1.18}}} \\
\multicolumn{1}{l|}{Conn$\downarrow$} &
  0.81 &
  0.41 &
  \multicolumn{1}{c|}{\colorbox{oorange}{0.30}} &
  \colorbox{oorange}{0.30} &
  0.41 &
  0.31 &
  \colorbox{rred}{\underline{0.26}} \\ \hline
\multicolumn{8}{l}{\textbf{\textit{VideoMatte}} ($1920 \times 1080$)} \\ \hline
\multicolumn{1}{l|}{MAD$\downarrow$} &
  11.13 &
  6.57 &
  \multicolumn{1}{c|}{5.81} &
  \colorbox{oorange}{4.42} &
  8.00 &
  \colorbox{oorange}{4.42} &
  \colorbox{rred}{\underline{4.24}} \\
\multicolumn{1}{l|}{MSE$\downarrow$} &
  \multicolumn{1}{c}{5.54} &
  \multicolumn{1}{c}{1.93} &
  \multicolumn{1}{c|}{0.97} &
  \multicolumn{1}{c}{\colorbox{oorange}{0.39}} &
  \multicolumn{1}{c}{3.24} &
  \multicolumn{1}{c}{0.40} &
  \multicolumn{1}{c}{\colorbox{rred}{\underline{0.33}}} \\
\multicolumn{1}{l|}{Grad$\downarrow$} &
  \multicolumn{1}{c}{15.30} &
  \multicolumn{1}{c}{10.55} &
  \multicolumn{1}{c|}{9.65} &
  \multicolumn{1}{c}{5.12} &
  \multicolumn{1}{c}{23.75} &
  \multicolumn{1}{c}{\colorbox{oorange}{4.03}} &
  \multicolumn{1}{c}{\colorbox{rred}{\underline{4.00}}} \\
\multicolumn{1}{l|}{dtSSD$\downarrow$} &
  3.08 &
  1.90 &
  \multicolumn{1}{c|}{1.78} &
  1.39 &
  2.37 &
  \colorbox{oorange}{1.31} &
  \colorbox{rred}{\underline{1.19}} \\ \hline \hline
  \multicolumn{8}{l}{\textbf{\textit{YoutubeMatte}} ($512 \times 288$)} \\ \hline
\multicolumn{1}{l|}{MAD$\downarrow$} &
  \multicolumn{1}{c}{19.37} &
  \multicolumn{1}{c}{4.08} &
  \multicolumn{1}{c|}{3.36} &
  \multicolumn{1}{c}{-} &
  \multicolumn{1}{c}{\colorbox{oorange}{3.08}} &
  \multicolumn{1}{c}{3.54} &
  \multicolumn{1}{c}{\colorbox{rred}{\underline{2.72}}} \\
\multicolumn{1}{l|}{MSE$\downarrow$} &
  16.21 &
  1.97 &
  \multicolumn{1}{c|}{\colorbox{oorange}{1.04}} &
  - &
  1.29 &
  1.23 &
  \colorbox{rred}{\underline{1.01}} \\
\multicolumn{1}{c|}{Grad$\downarrow$} &
  2.05 &
  1.34 &
  \multicolumn{1}{c|}{\colorbox{oorange}{1.03}} &
  - &
  1.16 &
  1.10 &
  \colorbox{rred}{\underline{0.97}} \\
\multicolumn{1}{l|}{dtSSD$\downarrow$} &
  \multicolumn{1}{c}{2.79} &
  \multicolumn{1}{c}{1.81} &
  \multicolumn{1}{c|}{\colorbox{oorange}{1.62}} &
  \multicolumn{1}{c}{-} &
  \multicolumn{1}{c}{1.83} &
  \multicolumn{1}{c}{1.88} &
  \multicolumn{1}{c}{\colorbox{rred}{\underline{1.60}}} \\
\multicolumn{1}{l|}{Conn$\downarrow$} &
  2.68 &
  0.60 &
  \multicolumn{1}{c|}{0.50} &
  - &
  \colorbox{oorange}{0.41} &
  0.49 &
  \colorbox{rred}{\underline{0.39}} \\ \hline
\multicolumn{8}{l}{\textbf{\textit{YoutubeMatte}} ($1920 \times 1080$)} \\ \hline
\multicolumn{1}{l|}{MAD$\downarrow$} &
  15.29 &
  4.37 &
  \multicolumn{1}{c|}{3.58} &
  - &
  6.49 &
  \colorbox{oorange}{2.37} &
  \colorbox{rred}{\underline{1.99}} \\
\multicolumn{1}{l|}{MSE$\downarrow$} &
  \multicolumn{1}{c}{12.68} &
  \multicolumn{1}{c}{2.25} &
  \multicolumn{1}{c|}{1.23} &
  \multicolumn{1}{c}{-} &
  \multicolumn{1}{c}{4.58} &
  \multicolumn{1}{c}{\colorbox{oorange}{0.98}} &
  \multicolumn{1}{c}{\colorbox{rred}{\underline{0.71}}} \\
\multicolumn{1}{l|}{Grad$\downarrow$} &
  \multicolumn{1}{c}{\colorbox{oorange}{8.42}} &
  \multicolumn{1}{c}{15.1} &
  \multicolumn{1}{c|}{12.97} &
\multicolumn{1}{c}{-} &
  \multicolumn{1}{c}{29.78} &
  \multicolumn{1}{c}{\colorbox{rred}{\underline{7.69}}} &
  \multicolumn{1}{c}{{8.91}} \\
\multicolumn{1}{l|}{dtSSD$\downarrow$} &
  2.74 &
  2.28 &
  \multicolumn{1}{c|}{2.04} &
  - &
  2.41 &
  \colorbox{oorange}{1.77} &
  \colorbox{rred}{\underline{1.65}} \\ \hline
\end{tabular}
}
\end{center}
\vspace{-2mm}
\end{table*}

\subsection{Recurrent Refinement During Inference}
\label{subsec:infer_strategy}
The first-frame matte is predicted from the given first-frame segmentation mask, and its quality will affect the matte prediction for the subsequent frames. The sequential prediction in the memory-based paradigm enables recurrent refinement during inference. Leveraging this mechanism, we introduce an optional first-frame warm-up module for inference. Specifically, we repeat the first frame $n$ times, treating each repetition as the initial frame, and use only the $n^{th}$ alpha output as the first frame to initialize the alpha memory bank. This (1) enhances robustness against the given segmentation mask and (2) refines matting details in the first frame to achieve image-matting quality (see Fig.~\ref{fig:refine} and Fig.~\ref{fig:suppl_refine} in the appendix).

\section{Data}
\label{sec:data}

We briefly introduce our new training datasets and benchmarks for evaluation, including both synthetic and real-world. More details are provided in the appendix (Sec.~\ref{sec:dataset}).

\subsection{Training Datasets}
\label{subsec:new_train_data}

To address limitations in video matting datasets in both quality and quantity, we collect abundant green screen videos, process them with Adobe After Effects, and conduct manual selection to remove common artifacts also found in VideoMatte240K~\cite{lin2021bgm} (see Fig.~\ref{fig:dataset_issue}). Compared to VideoMatte240K, our dataset, \textit{VM800}, is \textit{(1)} twice as large, \textit{(2)} more diverse in terms of hairstyles, outfits, and motion, and \textit{(3)} higher in quality. Ablation studies (Table~\ref{tab:ablation}(b) and Sec.~\ref{suppl_subsec:new_data}) further demonstrate the advantages of our dataset.

\subsection{Synthetic Benchmark} 
\label{subsec:syn_benchmark}
The standard benchmark, VideoMatte~\cite{lin2021bgm}, derived from VideoMatte240K, includes only \textit{5} unique foreground videos, which is under representative. Additionally, their foregrounds lack sufficient boundary details, limiting their ability to discern matting precision in boundary regions. To create a more comprehensive benchmark, we compile \textit{32} distinct 1920 $\times$ 1080 green-screen foreground videos from YouTube, and process them similarly to our training dataset. Our benchmark, \textit{YouTubeMatte}, provides enhanced detail representation, as reflected by higher Grad~\cite{rhemann2009perceptually} values.

\subsection{Real-world Benchmark and Metric}
\label{subsec:real_benchmark}
Real-world benchmarks are essential to facilitate the practical use of video matting models.
Although real-world videos lack ground truth (GT) alpha mattes, we can generate frame-wise segmentation masks as GT for \textit{core} areas benefiting from the high capability of existing VOS methods.
Specifically, we select a subset of 25 real-world videos~\cite{lin2022rvm} (100 frames each) with high-quality core GT masks verified manually. 
MAD, MSE, and dtSSD~\cite{erofeev2015perceptually} are then calculated at the core region as core region metrics, representing semantic stability that is critical for visual perception.

\section{Experiments}
\label{sec:experiments}

\begin{figure*}[t]
\begin{center}
    \vspace{-2mm}
    \includegraphics[width=\linewidth]{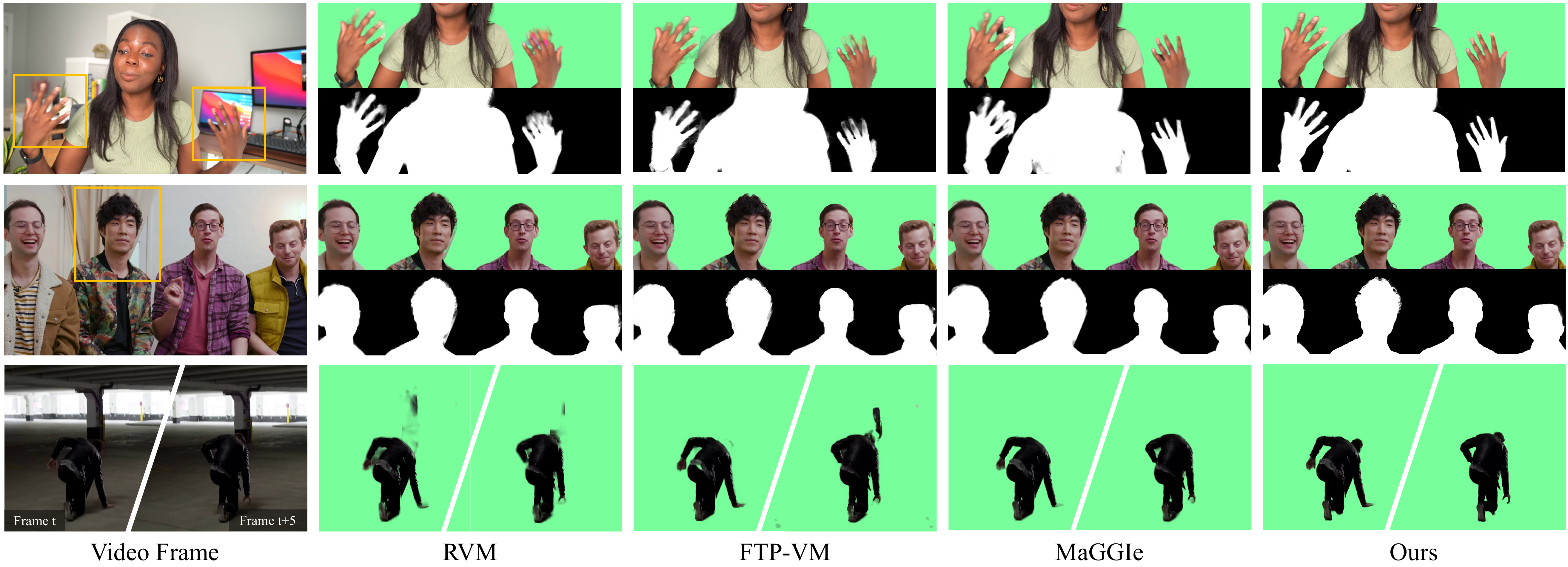}
    \vspace{-4mm}
    \caption{
    Qualitative comparisons on real-world videos.
    Our MatAnyone significantly outperforms existing auxiliary-free (RVM~\cite{lin2022rvm}) and mask-guided (FTP-VM~\cite{huang2023ftp} and MaGGIe~\cite{huynh2024maggie}) approaches in both detail extraction and semantic accuracy. For the lowest row, while other methods all miss out on important body parts (\ie, head) and mistakenly take background pixels as foreground (due to similar colors), thus generating messy outputs, our method presents an accurate and visually clean output by even identifying the shadow near the boundary.
    }
    \vspace{-4mm}
\label{fig:qualitative_real}
\end{center}
\end{figure*}
%
%\subsection{Training Schedule}

\noindent\textbf{Training Schedule.}
\textit{Stage 1.} 
Following the practice of RVM~\cite{lin2022rvm}, we start by training the entire model on our VM800 for 80k iterations.
The sequence length is initially set to 3 and extended to 8 with increasing sampling intervals for more complex scenarios.
\noindent \textit{Stage 2.} 
As the key stage, we apply the core supervision training strategy introduced in Section~\ref{subsec:cert_sup}.
Real segmentation data COCO~\cite{coco}, SPD~\cite{SPD} and YouTubeVIS~\cite{YouTubeVIS} are added for supervising the matting head.
The loss function applied are specified in Section~\ref{subsec:cert_sup}.
\noindent \textit{Stage 3.} 
Finally, we fine-tune the model with image matting data D646~\cite{qiao2020attention} and AIM~\cite{AIM} for finer matting details.

\subsection{Comparisons}
We compare MatAnyone with several state-of-the-art methods, including auxiliary-free (AF) methods: MODNet~\cite{ke2022MODNet}, RVM~\cite{lin2022rvm}, and RVM-Large~\cite{lin2022rvm}, and mask-guided methods: AdaM~\cite{lin2023adam}, FTP-VM~\cite{huang2023ftp}, and MaGGIe~\cite{huynh2024maggie}. 
\begin{table}[th]
% \vspace{-2mm}
\begin{center}
\setlength{\fboxsep}{3pt}
\caption{
    Quantitative comparisons on real-world benchmark~\cite{lin2022rvm}. The best and second performances are marked in \colorbox{rred}{\underline{red}} and \colorbox{oorange}{orange}, respectively.
    }
\label{tab:comparison_real}
\vspace{-2mm}
\renewcommand{\arraystretch}{1.1}
\renewcommand{\tabcolsep}{4.6mm}
\scalebox{0.84}{
\begin{tabular}{l|ccc}
\hline
Methods          & MAD$\downarrow$ & MSE$\downarrow$ & dtSSD$\downarrow$ \\ \hline
\multicolumn{4}{l}{{\textit{Auxiliary-free}}}  \\ \hline
MODNet~\cite{ke2022MODNet}    & 11.67 & 10.12 & 3.37  \\
RVM~\cite{lin2022rvm}      & 1.21  & 0.77  & 1.43  \\
RVM-Large~\cite{lin2022rvm}     & \colorbox{oorange}{0.95}  & \colorbox{oorange}{0.50}   & \colorbox{oorange}{1.30}   \\ \hline
\multicolumn{4}{l}{{\textit{Mask-guided}}}        \\ \hline
FTP-VM~\cite{huang2023ftp}     & 4.77  & 4.11  & 1.68  \\
MaGGIe~\cite{huynh2024maggie}      & 1.94  & 1.53  & 1.63  \\
\textbf{MatAnyone (Ours)} & \colorbox{rred}{\underline{0.14}}  & \colorbox{rred}{\underline{0.10}}  & \colorbox{rred}{\underline{0.89}}  \\ \hline
\end{tabular}
}
\end{center}
\vspace{-5mm}
\end{table}

\subsubsection{Quantitative Evaluations} 
\noindent\textbf{Synthetic Benchmarks.} 
For a comprehensive evaluation on synthetic benchmarks, we employ MAD (mean absolute difference) and MSE (mean squared error) for semantic accuracy, Grad (spatial gradient)~\cite{rhemann2009perceptually} for detail extraction, Conn (connectivity)~\cite{rhemann2009perceptually} for perceptual quality, and dtSSD~\cite{erofeev2015perceptually} for temporal coherence. In Table~\ref{tab:comparison_syn}, our method achieves the best MAD and dtSSD across all datasets at both high and low resolutions, demonstrating exceptional spatial accuracy for alpha mattes and remarkable temporal stability.
Apart from accuracy and stability, our method achieves the best Conn on both benchmarks, indicating its superior visual quality (Fig.~\ref{fig:qualitative_real} and Sec.~\ref{subsec:more_qua_comp} in the appendix).
\noindent\textbf{Real Benchmark.}
%
% For evaluations to be able to carry out on real benchmarks, we employ the metric for core region as discussed in Section~\ref{subsec:real_benchmark}. Table~\ref{tab:comparison_real} shows the superior generalizability of our method on real cases, by achieving the best metric values against both auxiliary-free and mask-guided methods by a considerable margin. 
For evaluation on real benchmarks, we use the core region metrics in Section~\ref{subsec:real_benchmark}. In Table~\ref{tab:comparison_real}, our method demonstrates superior generalizability on real cases, achieving the best metric values with a substantial margin over both auxiliary-free and mask-guided methods.

\subsubsection{Qualitative Evaluations}
\begin{figure*}[t]
\begin{center}
    \vspace{-3mm}
    \includegraphics[width=\linewidth]{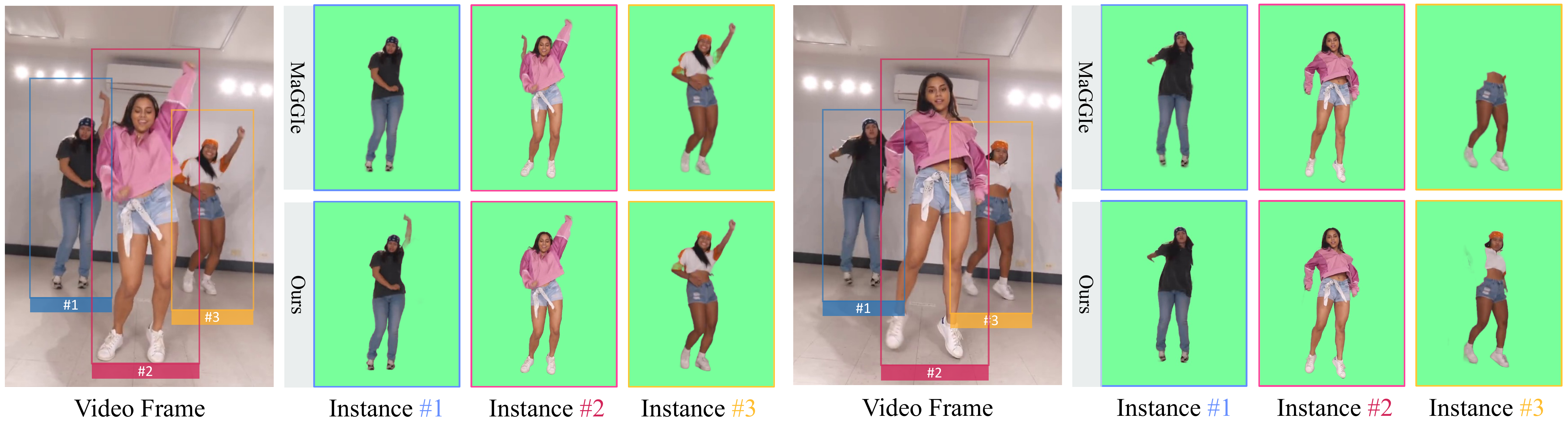}
    \vspace{-6mm}
    \caption{
    Quantitative comparisons with MaGGIe~\cite{huynh2024maggie} on instance video matting. Despite MaGGIe using instance mask as guidance for each frame, our method shows better performance, achieving better stability in object tracking and finer alpha matte details.
    }
    \vspace{-5mm}
\label{fig:qualitative_instance}
\end{center}
\end{figure*}

Visual results on real-world videos are in Fig.~\ref{fig:qualitative_real} and Fig.~\ref{fig:qualitative_instance}.

\noindent \textbf{General Video Matting.} MatAnyone outperforms existing auxiliary-free and mask-guided approaches in both detail extraction (\textit{boundary}) and semantic accuracy (\textit{core}).
%
% Specifically, MatAnyone is capable of generating more fine-grained details (\ie, hair) in Fig.~\ref{fig:qualitative_real}, and manages to distinguish the full human body from the extremely ambiguous background, where foreground and background colors are similar.
%
Fig.~\ref{fig:qualitative_real} shows that MatAnyone excels at fine-grained details (\eg, hair in the middle row) and differentiates full human body against complicated or ambiguous backgrounds when foreground and background colors are similar (\eg, last row).
\noindent \textbf{Instance Video Matting.} The assignment of target object at the first frame gives us flexibility for instance video matting. In Fig.~\ref{fig:qualitative_instance}, although MaGGIe~\cite{huynh2024maggie} benefits from using instance masks as guidance for \textit{each} frame, our method demonstrates superior performance in instance video matting, particularly in maintaining object tracking stability and preserving fine-grained details of alpha mattes.

% \noindent\textbf{Efficiency Comparison.}
% %
% %
% Table~\ref{tab: xx} presents the efficiency comparisons between all methods in terms of FLOPs and running time, which are computed based on a temporal length of 100.
% \shangchen{we can move this to suppl.}

\begin{figure*}[th]
\begin{center}
    \includegraphics[width=\linewidth]{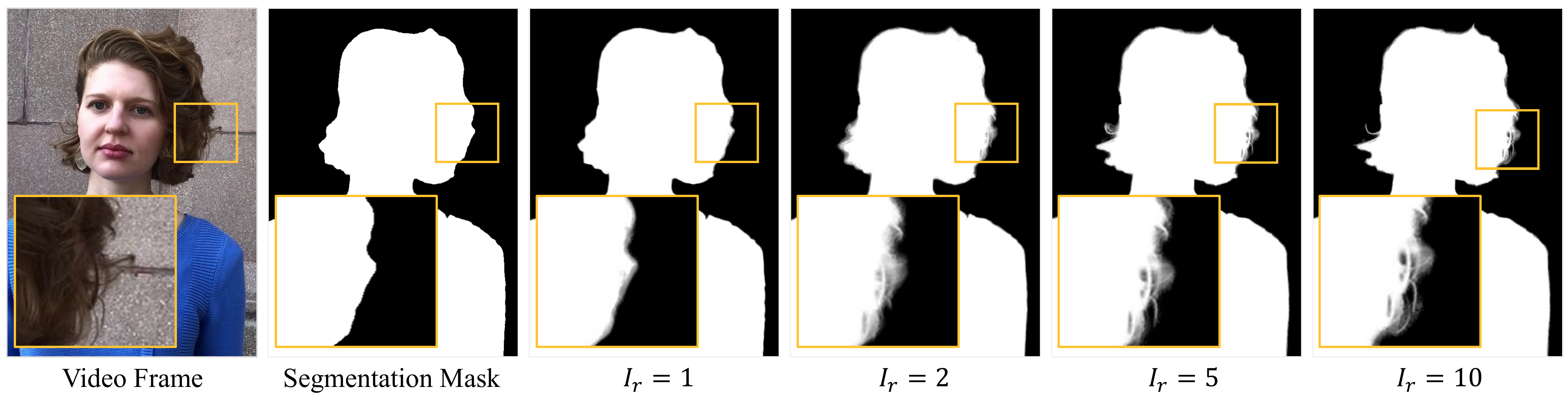}
    \vspace{-6mm}
    \caption{
    Improvement with Recurrent refinement. \textbf{(Zoom-in for best view)}
    }
    \vspace{-5mm}
\label{fig:refine}
\end{center}
\end{figure*}
\begin{figure}[t]
\begin{center}
    \includegraphics[width=\linewidth]{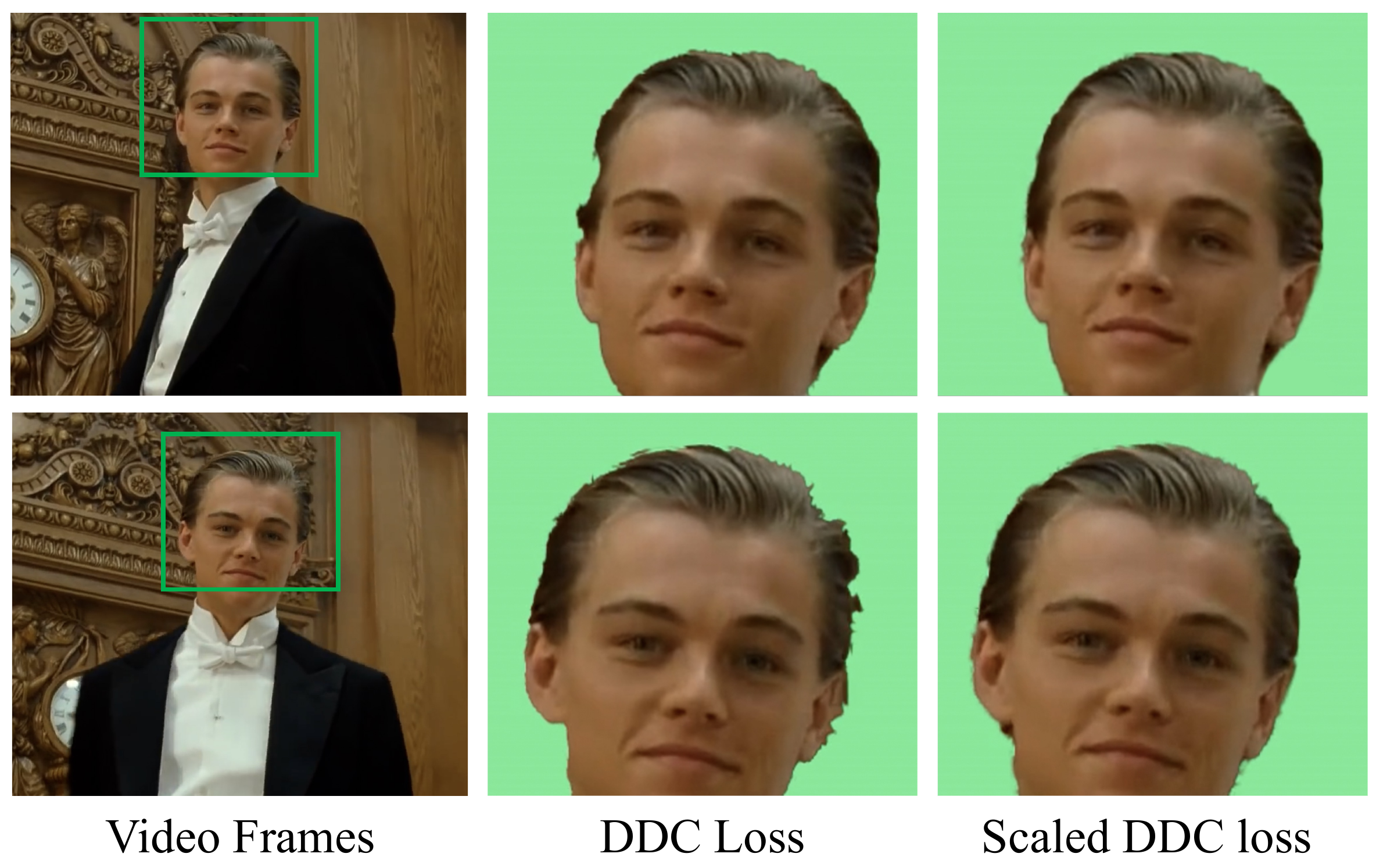}
    \vspace{-6mm}
    \caption{
    Comparison of matting results training with original DDC loss~\cite{liu2024ddc} and with scaled DDC loss, where the latter gives more stable and natural matting results.
    }
    \vspace{-8mm}
\label{fig:ddc}
\end{center}
\end{figure}
%
%

% \noindent\textbf{User Study.} real videos to show our advantages on real-world cases
% \shangchen{we can move this to suppl.}

\subsection{Ablation Study}
\label{sec:ablation}
\begin{table}[t]
\caption{Ablation study of the new training dataset (New Data), consistent memory propagation module (CMP), and new training scheme (New Training) on real benchmark (about $1080$p).
%
% For simplicity, models in the table are trained with the initial stage without fine-tuning with image matting data.
% 
}
\centering
\vspace{-2mm}
\renewcommand{\arraystretch}{1.1}
\renewcommand{\tabcolsep}{2mm}
\resizebox{\linewidth}{!} {
\begin{tabular}{l|ccc|ccc}
\toprule
Exp. & New Data & CMP & New Training & MAD$\downarrow$ & MSE$\downarrow$ & dtSSD$\downarrow$\\ 
\midrule
(a)   &   & & & 3.16 & 2.65 & 1.37\\
(b)  & \checkmark &   &    &  2.55  &  2.25 & 1.36 \\ 
(c) & \checkmark & \checkmark  &    & 1.85    &  1.67  & 1.25  \\ 
(d) &\checkmark & \checkmark  & \checkmark  &  {\bf 0.42} & {\bf 0.34} & {\bf 0.94}   \\ \bottomrule
\end{tabular}
}
\label{tab:ablation}
\vspace{-5mm}
\end{table}

\noindent\textbf{Enhancement from New Training Data.} 
In Table~\ref{tab:ablation}, by comparing (a) and (b), it is observed that training with new data noticeably improves the semantic performance with decreased MAD and MSE, showing that our newly-collected VM800 indeed contributes to robust training with its upgraded quantity, quality, and diversity.

\noindent\textbf{Effectiveness of Consistent Memory Propagation.} 
We further investigate the effectiveness of the consistent memory propagation (CMP) module. From Table~\ref{tab:ablation} (b) to (c), improvement can be seen across all metrics with CMP added, indicating its effectiveness in improving semantic stability and temporal coherency.
In particular, dtSSD in (c) is already lower than all the other methods in Table~\ref{tab:comparison_real}, showing the superiority of CMP in terms of temporal consistency.

\noindent\textbf{Effectiveness of New Training Scheme.} 
Our new training scheme brings our model to the next level with a noticeable improvement in all metrics. It already outperforms all the other methods in Table~\ref{tab:comparison_real} without further fine-tuning.
\noindent\textbf{Scaled DDC Loss.} 
We examine the merit of the scaled version of DDC loss by training with $\mathcal{L}_{core}$ and $\mathcal{L}_{boundary}$ only to maximize its effect. In Fig.~\ref{fig:ddc}, training with vanilla DDC loss produces segmentation-like jaggedness, especially among the boundary region. Our scaled DDC loss yields more stable and natural matting results.

\noindent\textbf{Effectiveness of Recurrent Refinement.}
Fig.~\ref{fig:refine} shows the effectiveness of recurrent refinement in a progressive manner. Given a rough segmentation mask, our method can produce alpha matte with descent details within 10 iterations.
\section{Conclusion}
\vspace{1mm}
\label{sec:conclusion}
We introduce MatAnyone, a practical framework for target-assigned human video matting that ensures stable and accurate results across diverse real-world scenarios. Our method leverages a region-adaptive memory fusion approach, which combines memory from previous frames to maintain semantic consistency in \textit{core} areas while preserving fine details along object \textit{boundaries}.  With a new training dataset that is larger, high-quality, and diverse and a novel training strategy that effectively leverages segmentation data, MatAnyone achieves robust and stable matting performance, even with complex backgrounds. These advancements position MatAnyone a practical solution for real-world video matting, also setting a solid foundation for future research in memory-based video processing.

\clearpage

\vspace{2mm}
\noindent{\bf Acknowledgement.} This study is supported under the RIE2020 Industry Alignment Fund – Industry Collaboration Projects (IAF-ICP) Funding Initiative, as well as cash and in-kind contribution from the industry partner(s).

{
    \small
    \bibliographystyle{ieeenat_fullname}
    \bibliography{main}
}

% CVPR 2024 Paper Template; see https://github.com/cvpr-org/author-kit
\clearpage
\renewcommand\thesection{\Alph{section}}
\onecolumn

\begin{center}
%	\vspace{-2mm}
	\Large\textbf{{Appendix}}\\
	\vspace{8mm}
\end{center}

In this supplementary material, we provide additional discussions and results to supplement the main paper. 
In Section~\ref{sec:arch}, we present the network details of our MatAnyone.
In Section~\ref{sec:training}, we discuss more training details, including training schedules, training augmentations, and loss functions.
In Section~\ref{sec:dataset}, we provide more details on our new training and testing datasets, including the generation pipeline and some examples for demonstration.
We present comprehensive results in Section~\ref{sec:more_results} to further show our performance, including those for ablation studies and qualitative comparisons. It is noteworthy that we also include a {\colorbox{oorange}{demo video (Section~\ref{subsec:demo_video})}} to showcase a Hugging Face demo and additional results on real-world cases in video format.

{   
    \hypersetup{linkcolor=blue}
    \tableofcontents
}

%%%%%%%%%%%%%%%%

% In this supplementary materials, we provide additional details, further discussions, and more results to supplement the main paper. In Sec.~\ref{sec:arch}, we present the architecture details of our proposed Upscale-A-Video.
% %
% In Sec.~\ref{sec:discussion}, we provide in-depth analysis on of the performance improvement achieved by our method and highlight its advantages.
% %
% Sec.~\ref{sec:result} contains more quantitative evaluations and visual comparisons.

\clearpage
\section{Architecture}
\label{sec:arch}
\subsection{Network Designs}
As illustrated in \cref{fig:overview} in the main paper, our MatAnyone mainly has five important components: (1) an \textit{encoder} for key and query transformation, (2) a \textit{consistent memory propagation} module for pixel memory readout, (3) an object transformer~\cite{cheng2024cuite} for memory grouping by object-level semantics, (4) a \textit{decoder} for alpha matte decoding, (5) a \textit{value encoder} for alpha matte encoding, which is used to update the alpha memory bank.
\noindent \textbf{Encoder.} 
We adopt ResNet-50~\cite{he2016resnet} for encoder following common practices in memory-based VOS~\cite{cheng2024cuite,cheng2021stcn,cheng2022xmem}. Discarding the last convolution stage, we take $\times 16$ downsampled feature as $F^t$ for key and query transformation, while features at scales $\times 8$, $\times 4$, $\times 2$, and $\times 1$ are used as skip connections for the decoder.
\noindent \textbf{Consistent Memory Propagation.} 
The process of consistent memory propagation is detailed in \cref{fig:overview}(b) in the main paper. \textit{Alpha memory bank} serves as the main working memory for past information query as in~\cite{cheng2022xmem,cheng2024cuite}, which is updated every $r^{th}$ frame across the whole time span. 
The query of the current frame to the alpha memory bank is implemented in an \textit{attention} manner following~\cite{cheng2022xmem,cheng2024cuite}. For the query $Q^{HW \times C}$ \footnote{We ignore the subscript $t$ in $Q_t$ for simplicity} and alpha memory bank $K^{THW \times C}$, $V^{THW \times C^v}$ \footnote{We ignore the subscript $m$ in $K_m$ and $V_m$ for simplicity}, the affinity matrix $A \in [0, 1]^{HW \times THW}$ of the query to alpha memory is computed as:
\begin{equation}
    A_{ij} = \frac{exp(d(Q_{i},K_{j}))}{\sum_z exp(d(Q_i, K_z))},
\end{equation}
where $d(\cdot,\cdot)$ is the anisotropic L2 function, $H$ and $W$ are the height and width at $\times 16$ downsampled input scale, and $T$ is the number of memory frames stored in alpha memory bank. The queried values $V_t^m$ in \cref{fig:overview}(b) in the main manuscript is obtained as:
\begin{equation}
    V^m_t = AV_m.
\end{equation}
In addition to that, we also maintain \textit{last frame memory} solely for the uncertainty prediction module we propose, and it is updated every frame.
The \textit{boundary-area prediction} module is lightweight with one $1 \times\ 1$ convolution and two $3 \times\ 3$ convolutions. By taking the input of a concatenation of current frame feature $K_t$, last frame feature $K_{t-1}$, and last alpha matte prediction $M_{t-1}$, it outputs a one-channel change probability mask $U_t$ of each query token, where higher $U_t$ indicates such token is likely to change more in the alpha value compared with $M_{t-1}$.
As mentioned in \cref{subsec:mem_prop} in the manuscript, the ground truth $U_t$ label is obtained by: $U^{GT}_t: |M^{GT}_{t-1} - M^{GT}_{t}| >= \delta $, where $\delta$ is set at 0 for segmentation data, and 0.001 for matting data as noise tolerance.
Since $U_t$ is predicted at a $\times 16$ downsampled scale in the memory space, the ground truth mask $U^{GT}_t$ is also downsampled in the mode of \texttt{area}.

\noindent \textbf{Object Transformer.} 
Our object transformer is derived from Cutie~\cite{cheng2024cuite} with three consecutive object transformer blocks. Pixel memory readout $P^t$ obtained from the consistent memory propagation module is then grouped through several attention layers and feed-forward networks. In this way, the noise brought by low-level pixel matching could be effectively reduced for a more robust matching against distractors. We do not claim contributions for this module.

\noindent \textbf{Decoder.} 
Our decoder is inspired by common practices in VOS~\cite{cheng2022xmem,cheng2024cuite} with modified designs specifically for the matting tasks. The mask decoder is VOS generally consists of two interactive upsampling from $\times 16$ to $\times 4$, and then a bilinear interpolation is applied to the input scale. However, since the boundary region for an alpha matte requires much more precision than a segmentation mask, we enrich the decoder with two more upsampling layers until $\times 1$, where skip connections from the encoder are applied at each scale to enhance the boundary precision.

\noindent \textbf{Value Encoder.} 
Similar to the encoder, we adopt ResNet-18~\cite{he2016resnet} for value encoder following common practices in memory-based VOS~\cite{cheng2024cuite,cheng2021stcn,cheng2022xmem}. Different from the encoder for key and query, the value encoder takes the predicted alpha matte $M^{t}$ as well as the image features as input, the encoded values are then used to update the alpha memory bank and last frame memory according to their updating rules.

\section{Training}
\label{sec:training}
\begin{table}[h]
\caption{
\textbf{Training settings and losses used in different training stages.}
$\dag$ indicates that segmentation loss is computed as an auxiliary loss on a \textit{segmentation} head, which will be abandoned during inference.
Other than that, matting loss and core supervision loss are computed on the \textit{matting} head for semantic stability in core regions and matting details in the boundary region.
}
\centering
\vspace{-2mm}
\renewcommand{\arraystretch}{1.15}
\renewcommand{\tabcolsep}{1.5mm}
\resizebox{\linewidth}{!} {
\begin{tabular}{l|cccc|ccc}
\toprule
Training Stage & \#Iterations & Matting Data  & Segmentation Data & Sequence Length & Matting Loss & Segmentation Loss$^\dag$ & Core Supervision Loss\\ 
\midrule
Stage 1  & 85K & video & image \& video & 3 (80K) $\rightarrow$ 8 (5K)& \checkmark & \checkmark  &     \\ 
Stage 2 & 40K & video & image \& video & 8 & \checkmark & \checkmark  &  \checkmark  \\ 
Stage 3 & 5K & image & image \& video & 8 & \checkmark & \checkmark  & \checkmark \\ \bottomrule
\end{tabular}
}
\label{tab:loss}
\vspace{-3mm}
\end{table}

\subsection{Training Schedules}
\noindent \textbf{Stage 1.} To initialize our model on memory propagation learning, we train with our new video matting data \textit{VM800}, which is of larger scale, higher quality, and better diversity than VideoMatte240K~\cite{lin2021bgm}. We use the AdamW~\cite{loshchilov2017adamw} optimizer with a learning rate of $1 \times 10^{-4}$ with a weight decay 0.001. The batch size is set to 16. We train with a short sequence length of 3 for 80K first, and then we train with a longer sequence length of 8 for another 5K for more complex scenarios. Video and image segmentation data COCO~\cite{coco}, SPD~\cite{SPD} and YouTubeVIS~\cite{YouTubeVIS} are used to train the segmentation head parallel to the matting head at the same time, as previous practices~\cite{lin2022rvm,lin2023adam,huang2023ftp}.
\noindent \textbf{Stage 2.} We apply our key training strategy - \textit{core-area supervision} in this stage. On the basis of the previous stage, we add additional supervision on the matting head with segmentation data to enhance the semantics robustness and generalizability towards real cases. In this stage, the learning rate is set to be $1 \times 10^{-5}$, and we train with a sequence length of 8 for 40K for both matting and segmentation data.
\noindent \textbf{Stage 3.} Due to the inferior quality of video matting data compared with image matting data annotated by humans, we finetune our model with image matting data instead for 5K with a $1 \times 10^{-6}$ learning rate. Noticeable improvements in matting details, especially among boundary regions, could be seen after this stage.
\subsection{Training Augmentations}
\noindent \textbf{Augmentations for Training Data.}
As discussed in the manuscript, video matting data are deficient in quantity and diversity.
In order to enhance training data variety during the composition process, we follow RVM~\cite{lin2022rvm} to apply motion (\eg, affine translation, scale, rotation, etc.) and temporal (\eg, clip reversal, speed changes, etc.) augmentations to both foreground and background videos.
Motion augmentations applied to image data also serve to synthesize video sequences from images, making it possible to fine-tune with higher-quality image data for details.

\noindent \textbf{Augmentations for Given Mask.}
Since our setting is to receive the segmentation mask for the first frame and make alpha matte prediction for all the frames including the first one, it is important to have our model robust to the given mask.
To generate the given mask in the training pipeline, we first obtain the original given mask. For segmentation data, it is just the ground truth (GT) for the first frame, while for matting data, it is the binarization result on the first-frame GT alpha matte, with a threshold of 50.
\textit{Erosion} or \textit{dilation} is then applied with a probability of 40\% each, with kernel sizes ranging from 1 to 5.
In this way, we force the model to learn alpha predictions based on an inaccurate segmentation mask, also enhancing the model robustness towards memory readout if it is not so accurate during the predictions in following frames.

\noindent \textbf{Augmentations for Assigned Object(s).}
The assignment of target object(s) as a segmentation mask for the first frame gives us flexibility for \textit{instance video matting}. Given the strong prior, the model is still easy to be confused by other salient humans not assigned as target.
To solve this, we find that a small modification in the video segmentation data pipeline has an obvious effect.
In YouTubeVIS~\cite{YouTubeVIS}, for each video with human existence, suppose the number of human instances is $H$. Instead of combining all of them as one object (practice in previous auxiliary-free methods~\cite{lin2022rvm}), we randomly take $h \leq H$ instance as foreground, while unchosen instances are marked as background.
In this way, we force the model to distinguish the target human object(s) even when other salient human object(s) exist, enhancing the robustness in object tracking for instance video matting even without instance mask for each frame as MaGGIe~\cite{huynh2024maggie} has.

\subsection{Loss Functions}
Given that we take the first-frame segmentation mask alongside with input frames as input, our model needs to predict alpha matte starting from the first frame, which is different from VOS methods~\cite{cheng2022xmem,cheng2024cuite}.
In addition, since we also apply mask augmentation on the given segmentation mask, the prediction from the segmentation head should also start from the first frame.
As a result, we need to apply losses on all $t \in [0, N]$ frames for both matting and segmentation heads.
There are mainly three kinds of losses involved in our training: (1) matting loss $\mathcal{L}^{mat}$; (2) segmentation loss $\mathcal{L}^{seg}$; (3) core supervision (CS) loss $\mathcal{L}^{cs}$, and their usages in different training stages are summarized in Table~\ref{tab:loss}. 
\noindent \textbf{Matting Loss.} For frame $t$, suppose we have the predicted alpha matte $M_t$ w.r.t. its ground-truth (GT) $M_t^{GT}$. We follow RVM~\cite{lin2022rvm} to employ L1 loss for semantics $\mathcal{L}_{l1}$, pyramid Laplacian loss~\cite{hou2019context} for matting details $\mathcal{L}_{lap}$, and temporal coherence loss~\cite{sun2021deep} $\mathcal{L}_{tc}$ for flickering reduction:
\begin{equation}
    \mathcal{L}_{l1} = \| M_t - M_t^{GT} \|_1,
\end{equation}
\begin{equation}
    \mathcal{L}_{lap} = \sum\limits^5_{s=1} \frac{2^{s-1}}{5} \| L_{pyr}^s(M_t) - L_{pyr}^s(M_t^{GT}) \|_1,
\end{equation}
\begin{equation}
    \mathcal{L}_{tc} = \| \frac{\text{d}M_t}{\text{d}t} - \frac{\text{d}M_t^{GT}}{\text{d}t} \|_2,
\end{equation}
The overall matting loss is summarized as:
\begin{equation}
    \mathcal{L}^{mat}=\mathcal{L}_{l1}+5\mathcal{L}_{lap}+\mathcal{L}_{tc}.
\end{equation}

\noindent \textbf{Segmentation Loss.} For frame $t$, suppose we have the predicted segmentation mask $S_t$ w.r.t. its ground-truth (GT) $S_t^{GT}$ from the segmentation head. We employ common losses used in VOS~\cite{cheng2022xmem,cheng2024cuite,yang2021dice}, $\mathcal{L}_{ce}$ and $\mathcal{L}_{dice}$. 
\begin{equation}
    \mathcal{L}_{ce} = S_t^{GT} (-log(S_t)) + (1 - S_t^{GT}) (-log(1 - S_t)),
\end{equation}
\begin{equation}
    \mathcal{L}_{dice} = 1 - \frac{2 S_t S_t^{GT}+1}{S_t+S_t^{GT}+1}.
\end{equation}
The overall segmentation loss is summarized as:
\begin{equation}
    \mathcal{L}^{seg}=\mathcal{L}_{ce}+\mathcal{L}_{dice}.
\end{equation}
\noindent \textbf{Core Supervision Loss.} For core-area supervision, we combine the region-specific losses: $\mathcal{L}_{core}$ for core region and $\mathcal{L}_{boundary}$ for boundary region as defined in \cref{subsec:cert_sup} in the manuscript, and the overall core supervision loss is summarized as:
\begin{equation}
    \mathcal{L}^{cs}=\mathcal{L}_{core}+1.5\mathcal{L}_{boundary}.
\end{equation}

\section{Dataset}
\label{sec:dataset}
\begin{table}[h]
\caption{
\textbf{Comparison on Datasets.}
We compare our new training data and testing data with the old ones, in terms of the number of distinct foregrounds, sources, and whether harmonization is applied.}
\centering
\vspace{-2mm}
\renewcommand{\arraystretch}{1.15}
\renewcommand{\tabcolsep}{1.5mm}
\resizebox{\linewidth}{!} {
\begin{tabular}{l|cc|cc}
\toprule
Datesets & VideoMatte240K (old train)~\cite{lin2021bgm} & \textbf{VM800} (new train) & VideoMatte (old test)~\cite{lin2021bgm} & \textbf{YouTubeMatte} (new test)\\ 
\midrule
\#Foregrounds  & 475 & 826 & 5 & 32   \\ 
Sources & - & Storyblocks, Envato Elements, Motion Array & - & YouTube \\ 
Harmonized & - & - & \text{\sffamily x} & \checkmark \\
\bottomrule
\end{tabular}
}
\label{tab:data}
\vspace{-3mm}
\end{table}

\subsection{New Training Dataset - VM800}
\label{subsec:suppl_new_data}
\noindent \textbf{Overview.} 
%
% compared with VideoMatte240k (table \& fig): quantity (table); quality (figure: details \& flaws); diversity (table \& figure)
%
As summarized in Table~\ref{tab:data}, our new training dataset \textit{VM800} has almost \textbf{twice} the number of foreground videos than VideoMatte240K~\cite{lin2021bgm} in quantity. 
To enhance diversity and data distribution, our foreground green screen videos are downloaded from a total of \textit{three} video footage websites: Storyblocks, Envato Elements, and Motion Array, and thus enjoy a diversity in hairstyles, outfits, and motion.
In addition, we ensure the high quality of our VM800 dataset in fine detail and through careful manual selection. 

\noindent \textbf{Generation Pipeline.}
We employ Adobe After Effects in our data generation pipeline to extract alpha channels from green screen footage videos.
Since the amount of green screen footage to be processed is huge, we would like to obtain the preliminary results with an automatic pipeline. 
We first use \texttt{Keylight} and set \texttt{Screen Color} to be the pixel value taken from the upper left corner for each frame. To obtain a clean alpha matte, we clip the values smaller than 20 to be 0 and those larger than 80 to be 255. To further enhance the alpha matte quality, we post-process with another two keying effects \texttt{Key Cleaner} and \texttt{Advanced Spill Supressor}, which are generally used together following \texttt{Keylight}. Since we are processing a video, we also turn on \texttt{reduce chatter} in \texttt{Key Cleaner} to reduce flickering in the boundary region.
For batch processing, we compile the above process into a Javascript and XML file for After Effects to run with, and obtain a large batch of preliminary results for manual selection.
{
\centering
\small
\begin{tcolorbox}[colframe=blue!50, colback=blue!1, boxrule=0.2mm, arc=2mm, width=0.8\textwidth]
{\color{green!65!black}{
% \noindent\rule{0.7\linewidth}{0.7pt}
\vspace{1mm}
\begin{verbatim}
Keylight
    - Screen Color: pixel value of upper left corner
    - Screen Matte:
           - Clip Black: 20
           - Clip White: 80
    
Key Cleaner
    - radius: 1
    - reduce chatter: check
    
Advanced Spill Supressor
\end{verbatim}
\vspace{-2.5mm}
}}
% \rule{0.7\linewidth}{0.7pt}
\end{tcolorbox}
}
\vspace{3mm}
\begin{figure*}[h]
\begin{center}
    % \vspace{-4mm}
    \includegraphics[width=\linewidth]{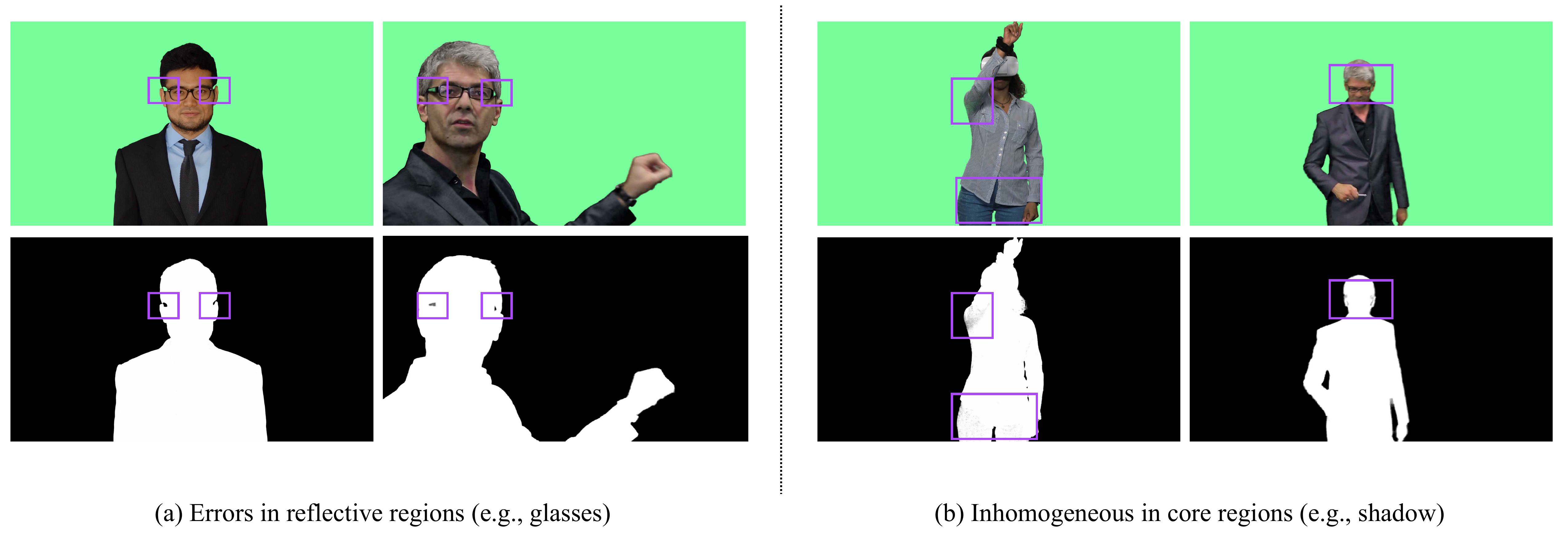}
    \vspace{-3mm}
    \caption{
    \textbf{Issues with VideoMatte240K~\cite{lin2021bgm}.}
    (a) Errors in alpha values exist in reflective regions (\eg, ``a hole" on glasses).
    (b) Inhomogeneous alpha values exist in core regions (\eg, caused by shadow), where the alpha value should be exactly 0 or 1.}
    \vspace{-4mm}
\label{fig:dataset_issue}
\end{center}
\end{figure*}

\begin{figure*}[h]
\begin{center}
    % \vspace{-4mm}
    \includegraphics[width=\linewidth]{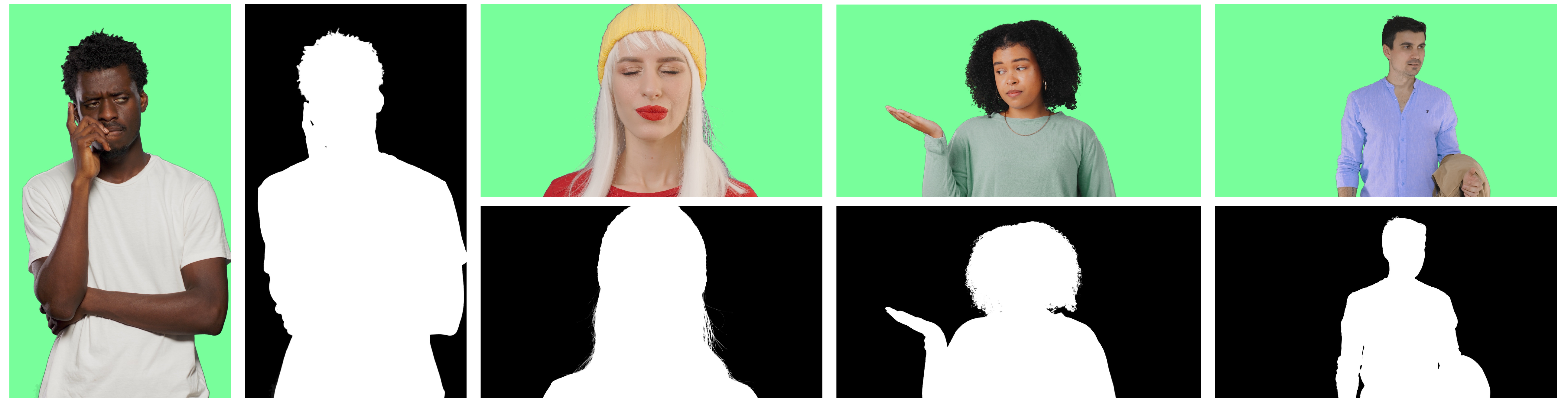}
    \vspace{-3mm}
    \caption{
    \textbf{Gallery for our new training dataset VM800.}
    High-quality details in the boundary regions and diversity in terms of gender, hairstyles, and aspect ratios could be clearly observed.
    }
    \vspace{-4mm}
\label{fig:dataset_compare}
\end{center}
\end{figure*}

\noindent \textbf{Quality - Fine Details.} The green screen foreground videos we downloaded are almost in a \textbf{4K} quality, and we also place a higher priority on those videos with more details (\eg, hair) in our download choice. Fig.~\ref{fig:dataset_compare} shows the fine details in our VM800 dataset.
\noindent \textbf{Quality - Careful Manual Selection.} We notice that alpha mattes extracted with After Effects from green screen videos often encounter inhomogeneities in core regions. For example, reflective regions in the foreground will result in a near-zero value (\ie, a hole) in the alpha matte, as shown in Fig.~\ref{fig:dataset_issue}(a). 
In addition, noise also exists in the green screen background, resulting in the fact the alpha values may not homogeneously equal 0, which should not be the case in the core region. Similarly, for foregrounds, colors that are similar to the background green, or shadow in the foreground, may also result in the alpha values not homogeneously equal to 1 in the core foreground region, making the alpha matte look noisy, as shown in Fig.~\ref{fig:dataset_issue}(b).
Since VideoMatte240K~\cite{lin2021bgm} is also obtained with After Effects, we observe that alpha mattes with the above problems still exist, and thus taking such wrong ground truth for training will inevitably lead to problematic inference results (Fig.~\ref{fig:new_trainig_data_scheme}(a)).
As a result, we conduct careful manual selection to examine all our processed alpha mattes, and leave out those with the above problems. As shown in Fig.~\ref{fig:new_trainig_data_scheme}(a), training with our VM800 will not lead to such problematic results.
\subsection{New Test Dataset - YouTubeMatte}
\begin{figure*}[h]
\begin{center}
    % \vspace{-4mm}
    \includegraphics[width=\linewidth]{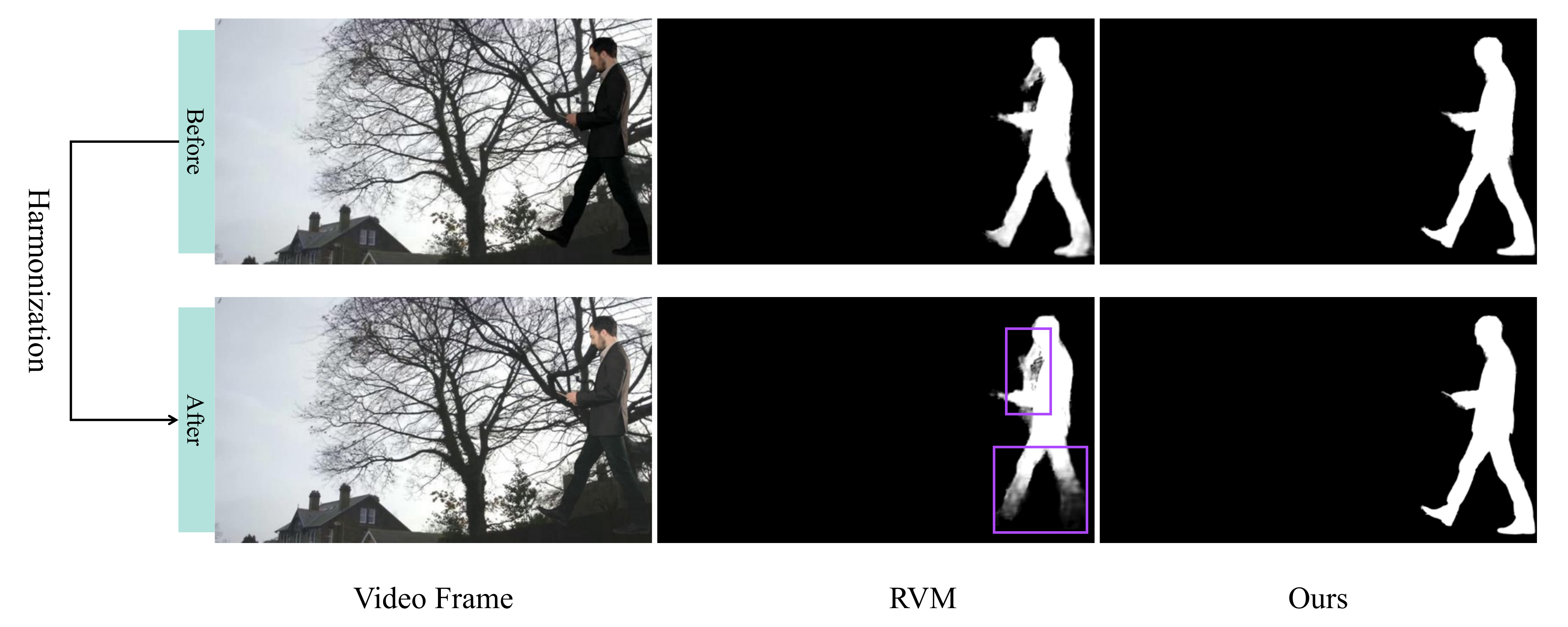}
    \vspace{-3mm}
    \caption{
    \textbf{Harmonization on synthetic benchmarks and its effect on model performance.}
    Harmonization~\cite{ke2022harmonizer} is an operation that makes the composited frame more natural and realistic, which also effectively makes our YouTubeMatte a more challenging benchmark that is closer to the real distribution.
    It is observed that while RVM~\cite{lin2022rvm} is confused by the harmonized frame, our method still yields robust performance.
    }
    \vspace{-4mm}
\label{fig:dataset_test_harmonization}
\end{center}
\end{figure*}

\noindent \textbf{Overview.}
As summarized in Table~\ref{tab:data}, our new synthetic benchmark \textit{YouTubeMatte} has over \textbf{six times} larger than the number of distinct foreground videos in VideoMatte~\cite{lin2021bgm}, making it a much more representative benchmark for evaluation with better diversity.
In addition, the green screen videos for foregrounds are downloaded from YouTube at a scale of $1920 \times 1080$ with rich boundary details, thus enhancing its ability to discern matting precision in boundary regions.
While the generation pipeline for YouTubeMatte is almost the same as that for VM800, \textbf{harmonization}~\cite{ke2022harmonizer}, however, is applied when compositing the foreground on a background. Such an operation effectively makes YouTubeMatte a more challenging benchmark that is closer to the real distribution.
As shown in Fig.~\ref{fig:dataset_test_harmonization}, while RVM~\cite{lin2022rvm} is confused by the harmonized frame, our method still yields robust performance.
%
% \begin{figure*}[h]
% \begin{center}
%     % \vspace{-4mm}
%     \includegraphics[width=\linewidth, height=5cm]{example-image-duck}
%     \vspace{-3mm}
%     \caption{
%     Diversity [show different categories]
%     }
%     \vspace{-4mm}
% \label{fig:dataset_test_overview}
% \end{center}
% \end{figure*}
%

\subsection{Real Benchmark and Evaluation}
\noindent \textbf{Overview.}
As a technique towards real-world applications (\eg, virtual background in the online meeting), the synthetic benchmark is not enough to test the generalizability of video matting models. 
Although there are countless of real human videos for testing in the wild, the lack of GT alpha mattes makes them hard to serve as a real benchmark. Here, we select a subset of 25 real-world videos from~\cite{lin2022rvm}, where a consecutive of 100 frames for each video are selected with no scene transition, to form our real benchmark.
According to our definitions in \cref{fig:motivation}(a) in the manuscript, we could also divide the evaluation metrics for core regions and for boundary separately, making evaluation for real benchmarks feasible.
%
% \begin{figure*}[h]
% \begin{center}
%     % \vspace{-4mm}
%     \includegraphics[width=\linewidth, height=4cm]{example-image-duck}
%     \vspace{-3mm}
%     \caption{
%     visualization of core region.
%     }
%     \vspace{-4mm}
% \label{fig:core_region}
% \end{center}
% \end{figure*}

\noindent \textbf{Evaluation on Core Regions.}
Thanks to the recent success of VOS methods~\cite{cheng2022xmem,cheng2024cuite}, frame-wise segmentation masks could be generated with high precision. Here, we employ Cutie~\cite{cheng2024cuite} for video segmentation results. We first obtain the trimap for each segmentation mask by applying dilation and erosion (with kernel size 21), and then compute the core mask where trimap values equal 0 or 1.
In this way, the values of a segmentation mask within its core region could be considered as the GT alpha values for the core region, where common metrics including MAD and MSE for semantic accuracy, and dtSSD~\cite{erofeev2015perceptually} for temporal coherency could be applied for evaluation.

\section{More Results}
\label{sec:more_results}
\subsection{Enhancement from New Training Data}
\label{suppl_subsec:new_data}
\begin{figure*}[h]
\begin{center}
    % \vspace{-4mm}
    \includegraphics[width=\linewidth]{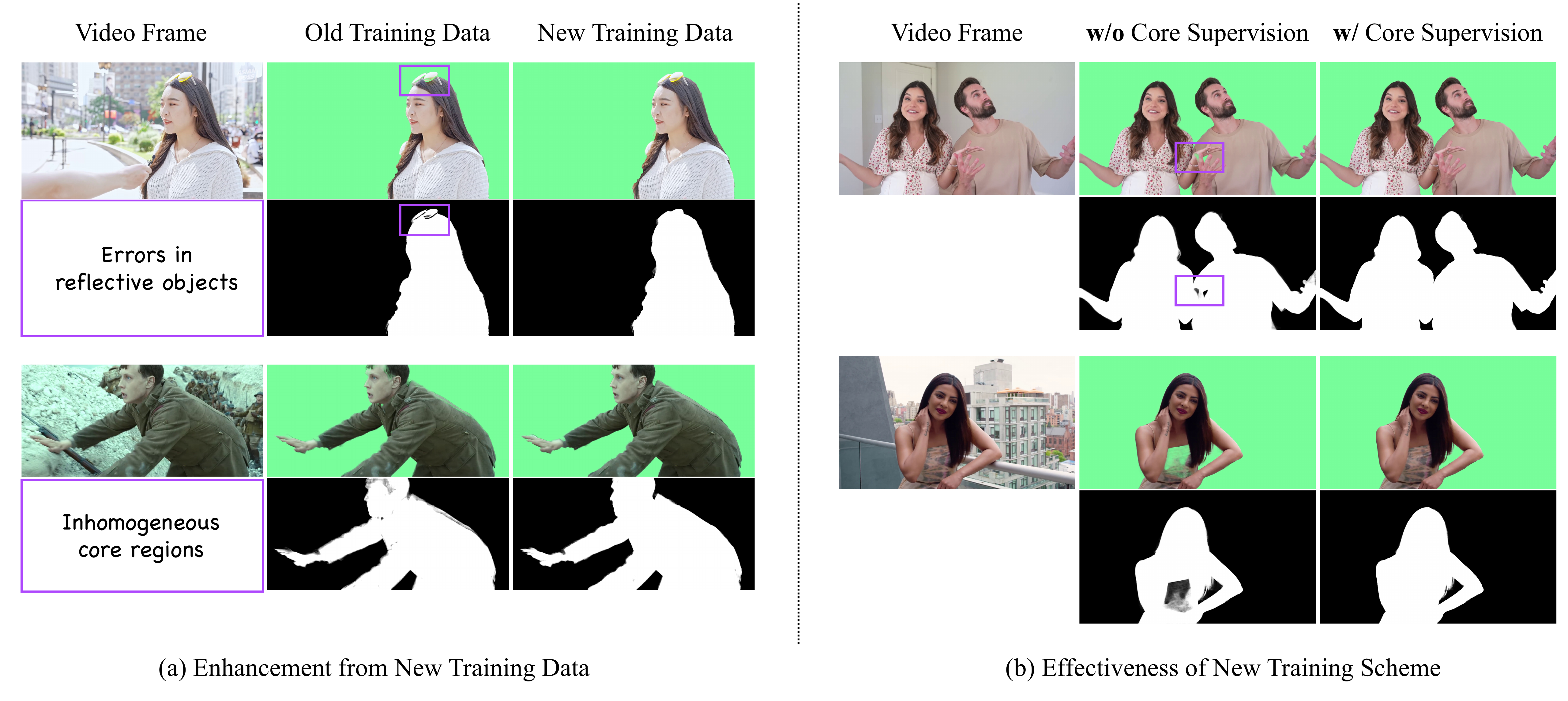}
    \vspace{-4mm}
    \caption{
    \textbf{(a) Comparison on results trained with old training data (VideoMatte240K~\cite{lin2021bgm}) and new training data (our VM800).} It could be observed that training with old data will lead to errors in reflective objects (\eg, holes on the sunglasses) and inhomogeneous alpha values in the core regions. However, both issues are fixed when training with our new data, indicating a higher quality.
    \textbf{(b) Comparison on results trained without and with core-area supervision.} It could be observed that training without it will lead to semantics error due to the weak supervision from real segmentation data, while training with core supervision largely improves semantics accuracy thanks to the stronger supervision enabled.
    }
    \vspace{-4mm}
\label{fig:new_trainig_data_scheme}
\end{center}
\end{figure*}

As discussed in \cref{subsec:new_train_data} in the manuscript and Section~\ref{subsec:suppl_new_data} in the supplementary, our new training data VM800 is upgraded in quantity, quality, and diversity. In addition to the quantitative evaluation in \cref{tab:ablation} in the manuscript, we further show the enhancement from new training data by providing more results when comparing the model trained with VideoMatte240K~\cite{lin2021bgm} and the model trained with our VM800 in Fig.~\ref{fig:new_trainig_data_scheme}(a).

\subsection{Effectiveness of Consistent Memory Propagation}
\label{suppl_subsec:cmp}
\begin{figure*}[h]
\begin{center}
    % \vspace{-4mm}
    \includegraphics[width=\linewidth]{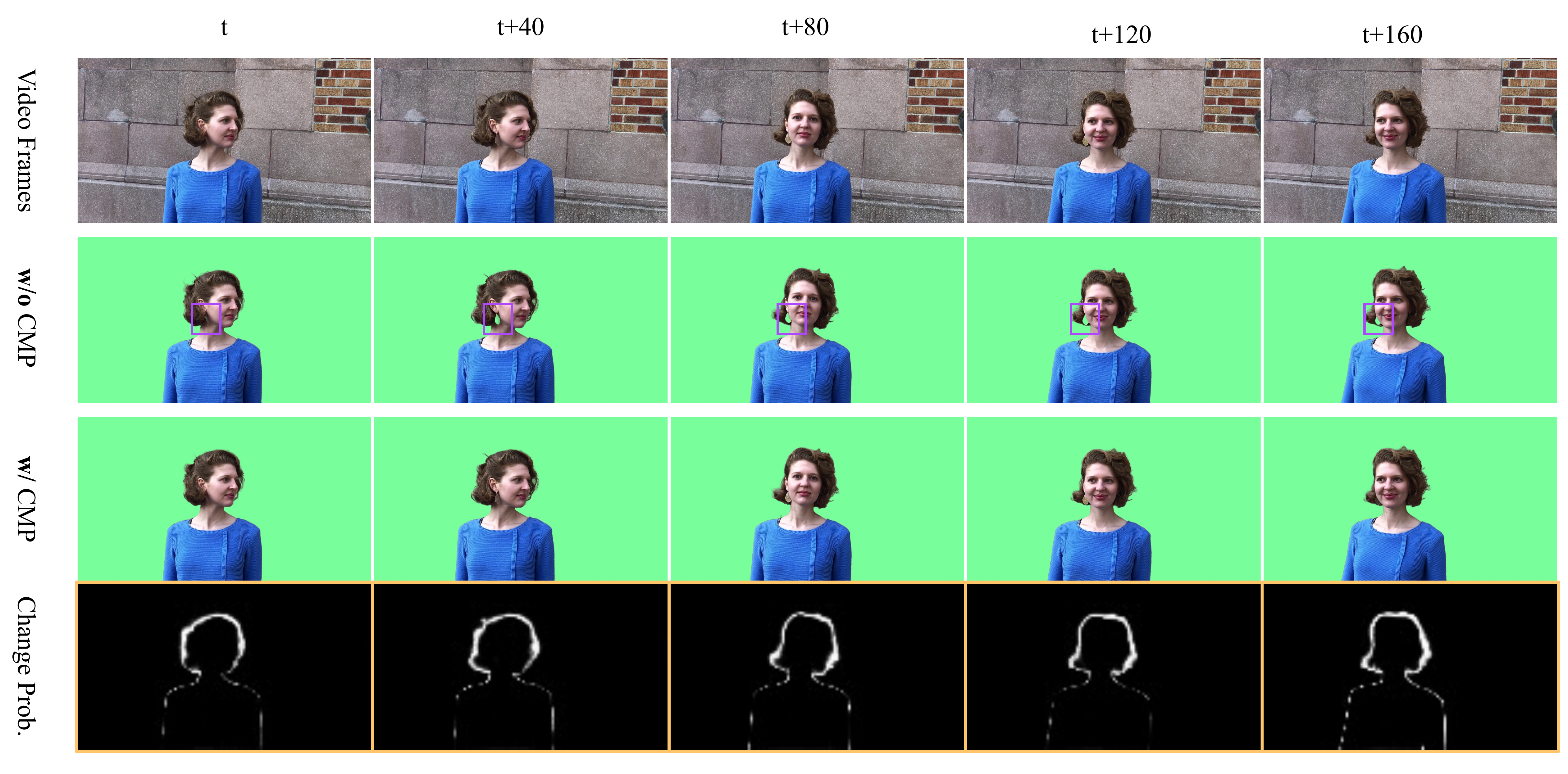}
    \vspace{-4mm}
    \caption{
    \textbf{Comparison on results with and without Consistent Memory Propagation.} It could be observed that when CMP is not applied, semantic errors constantly exist across a wide span of video frames. However, when training with CMP, we observe from the ``Change Probability" mask that usually our model only takes pixels near the boundary as ``changed", and most of the inner regions (\ie, earring) will mainly take the memory values from the last frame. As we can see on the figure, while predictions are both correct at time $t$, the model with CMP successfully keeps the correctness and gives stable results, while the model without CMP quickly breaks the correctness and never recovers.
    }
    \vspace{-4mm}
\label{fig:cmp}
\end{center}
\end{figure*}

As one of our key designs, the consistent memory propagation (CMP) module improves both stability in core regions and quality in boundary details. In addition to the quantitative evaluation in \cref{tab:ablation} in the manuscript, we give more qualitative results and analysis in Fig.~\ref{fig:cmp}.

\subsection{Effectiveness of New Training Scheme}
Our new training scheme introduces core-area supervision, which largely enhances the semantic accuracy and stability, as shown in \cref{tab:ablation} in the manuscript. More qualitative results are shown in Fig.~\ref{fig:new_trainig_data_scheme}(b) for better visualization of its effects.

\subsection{Effectiveness of Recurrent Refinement}
\begin{figure*}[h]
\begin{center}
    % \vspace{-4mm}
    \includegraphics[width=\linewidth]{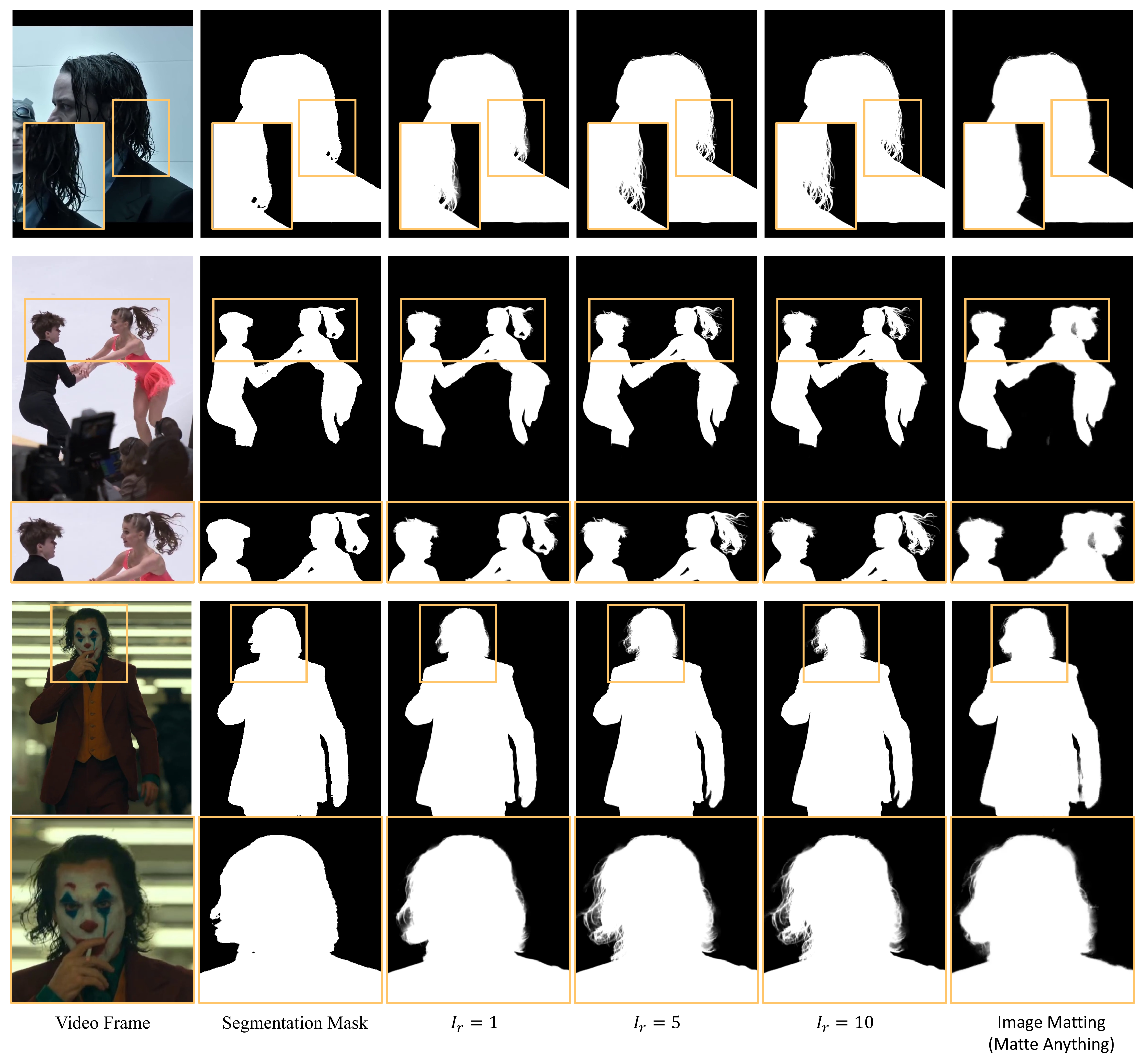}
    \vspace{-3mm}
    \caption{
    \textbf{Comparison on results with iterative refinement.} A noticeable enhancement on details can be observed even with one iteration of refinement compared with the given segmentation mask. Within 10 iterations, our model is able to achieve matting details at an image-matting level, even better than Matte Anything~\cite{yao2024matteanything}, which is an image matting model also based on the results from SAM~\cite{kirillov2023sam}.
    }
    \vspace{-4mm}
\label{fig:suppl_refine}
\end{center}
\end{figure*}

As discussed in \cref{subsec:infer_strategy} in the manuscript, the sequential prediction in the memory-based paradigm enables recurrent refinement without the need for retraining during inference. By repeating the first frame $n$ times and iteratively updating the first frame prediction based on the last-time prediction, the quality of the first frame alpha matte could be recurrently refined. We show in Fig.~\ref{fig:suppl_refine} that such recurrent refinement can not only (1) enhance the robustness to the given segmentation mask even when it is of low quality, but also (2) achieve matting details at an image-matting level when compared with an image matting method (\ie, Matte Anything~\cite{yao2024matteanything} in the last column).

\subsection{More Qualitative Comparisons}
\label{subsec:more_qua_comp}
\begin{figure*}[h]
\begin{center}
    % \vspace{-4mm}
    \includegraphics[width=\linewidth]{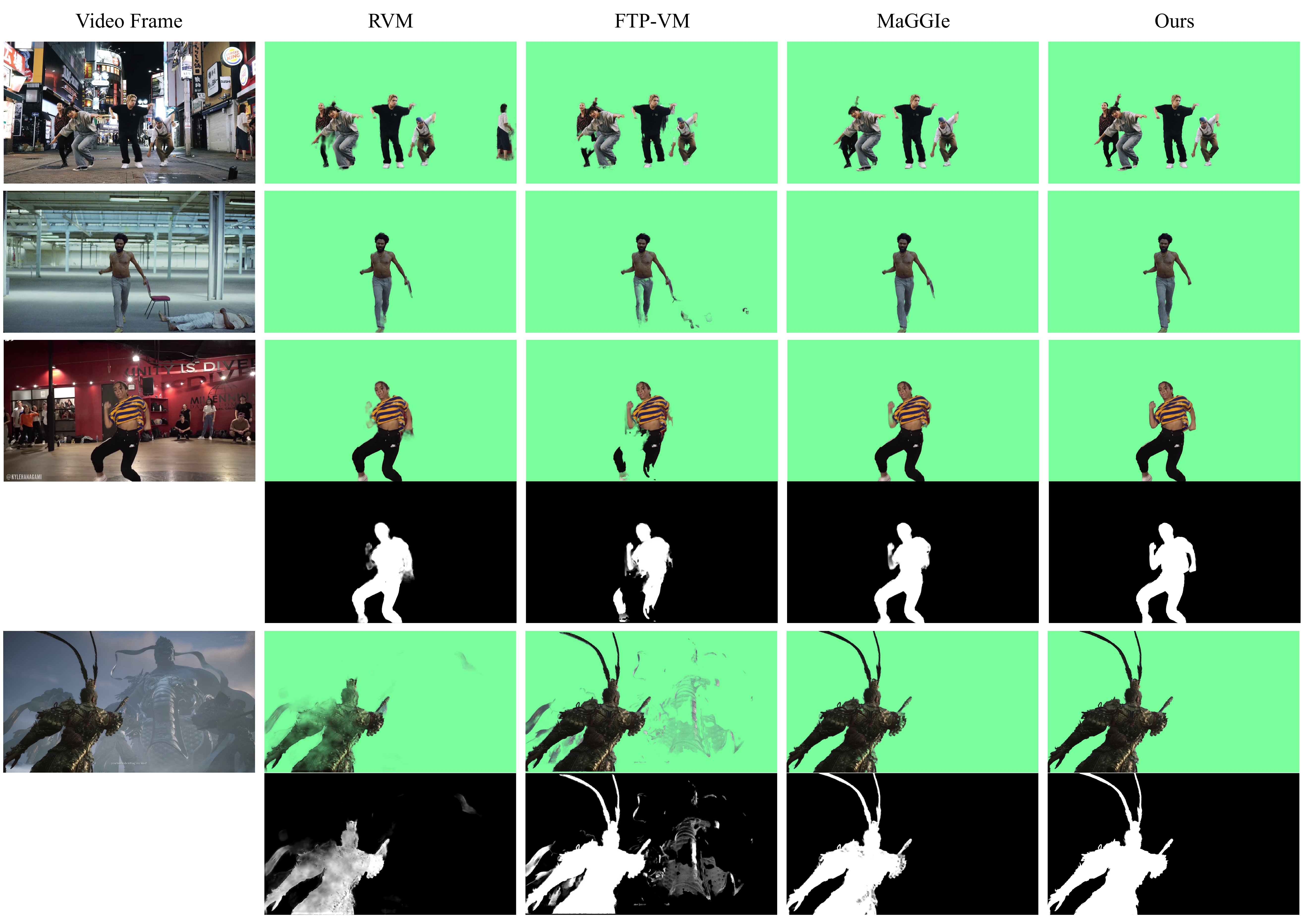}
    \vspace{-3mm}
    \caption{
    \textbf{More qualitative comparisons on general video matting with SOTA methods.} We compare our MatAnyone with both auxiliary-free (AF) method: RVM~\cite{lin2022rvm} and mask-guided methods: FTP-VM~\cite{huang2023ftp}, and MaGGIe~\cite{huynh2024maggie}.
    It could be observed that our method significantly outperforms others in both detail extraction and semantic accuracy, across diverse and complex real scenarios.
    It is noteworthy that although sometimes MaGGIe~\cite{huynh2024maggie} seems to give acceptable results when compositing with a green screen, its alpha matte turns out to be noisy (\ie, inhomogeneous in the core foreground region and blurry in the boundary region), while our alpha matte is clean with fine-grained details in the boundary region. As a result, we also include alpha mattes for a more comprehensive comparison. \textbf{(Zoom in for best view)}
    }
    \vspace{-4mm}
\label{fig:more_resutls_general}
\end{center}
\end{figure*}

\begin{figure*}[h]
\begin{center}
    % \vspace{-4mm}
    \includegraphics[width=\linewidth]{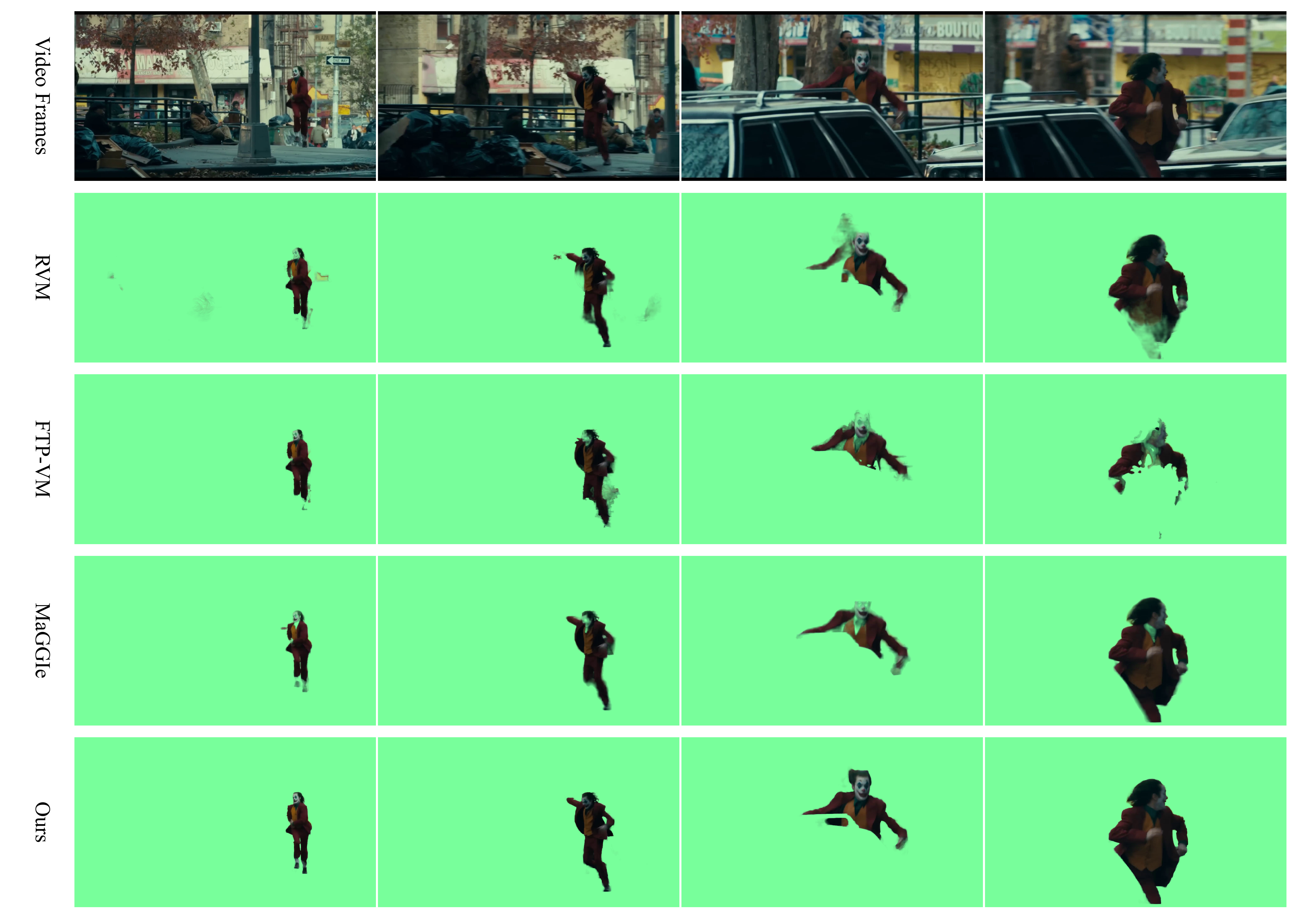}
    \vspace{-3mm}
    \caption{
    \textbf{A challenging example of general video matting across a long time span.} We compare our MatAnyone with both auxiliary-free (AF) method: RVM~\cite{lin2022rvm} and mask-guided methods: FTP-VM~\cite{huang2023ftp}, and MaGGIe~\cite{huynh2024maggie}.
    It could be observed that our model is able to track the target object stably even when the object is moving fast in a highly complex scene, where all the other methods present noticeable failures. \textbf{(Zoom in for best view)}
    }
    \vspace{-4mm}
\label{fig:more_resutls_general_long1}
\end{center}
\end{figure*}

\begin{figure*}[h]
\begin{center}
    % \vspace{-4mm}
    \includegraphics[width=\linewidth]{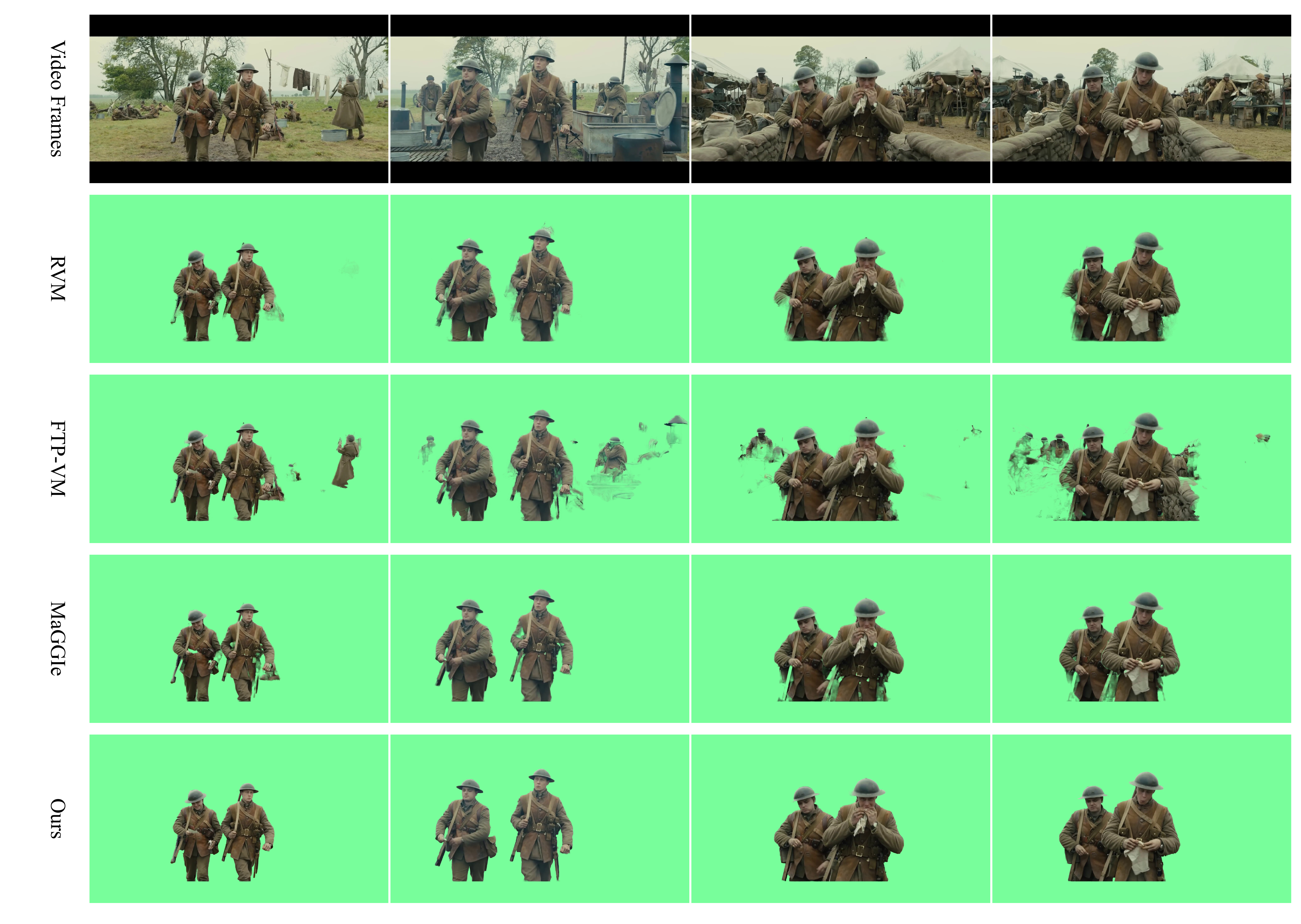}
    \vspace{-3mm}
    \caption{
    \textbf{Another challenging example of general video matting across a long time span.} We compare our MatAnyone with both auxiliary-free (AF) method: RVM~\cite{lin2022rvm} and mask-guided methods: FTP-VM~\cite{huang2023ftp}, and MaGGIe~\cite{huynh2024maggie}.
    This example showcases that our model is able to track the target objects even in a highly ambiguous background, where the colors for foreground and background are similar, and also multiple humans in the background.
    In addition, it also demonstrates when there is more than one target object, our model is still able to handle this challenging case well. \textbf{(Zoom in for best view)}
    }
    \vspace{-4mm}
\label{fig:more_resutls_general_long2}
\end{center}
\end{figure*}

\begin{figure*}[h]
\begin{center}
    % \vspace{-4mm}
    \includegraphics[width=\linewidth]{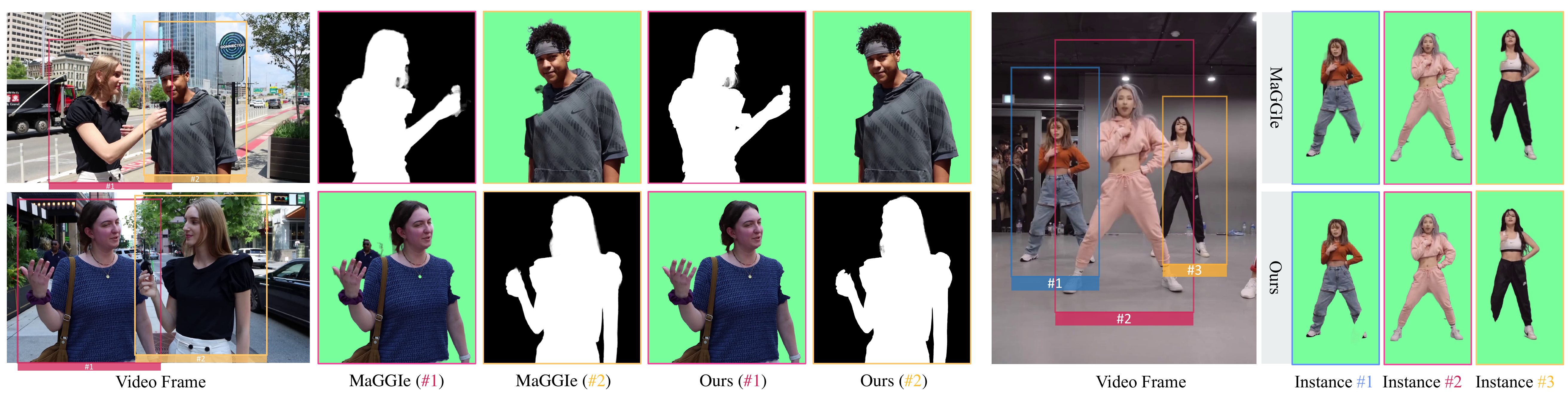}
    \vspace{-3mm}
    \caption{
    \textbf{More qualitative comparisons on instance matting.}
    We compare our MatAnyone with MaGGIe~\cite{huynh2024maggie}, a mask-guided method that requires the instance mask for \textit{each} frame, while our method only requires the mask for the \textit{first} frame.
    It could be observed that even with such strong given prior, MaGGIe still performs below our method in terms of semantic accuracy in the core regions. 
    Moreover, in terms of the boundary regions, by examining the details there, we could clearly observe that the details generated by MaGGIe are blurry and far from fine-grained compared with our results. \textbf{(Zoom in for best view)}
    }
    \vspace{-4mm}
\label{fig:more_resutls_instance}
\end{center}
\end{figure*}

In this subsection, we provide additional visual comparisons of our method with the state-of-the-art methods, including auxiliary-free (AF) method: RVM~\cite{lin2022rvm} and mask-guided methods: FTP-VM~\cite{huang2023ftp}, and MaGGIe~\cite{huynh2024maggie}.
Fig.~\ref{fig:more_resutls_general} presents the general video matting results on real videos. To further demonstrate the superiority of our model, Fig.~\ref{fig:more_resutls_general_long1} and Fig.~\ref{fig:more_resutls_general_long2} both showcase a challenging case respectively, where other methods mostly fail.
In addition, Fig.~\ref{fig:more_resutls_instance} demonstrates the instance matting results compared with MaGGIe~\cite{huynh2024maggie}, a method with instance mask for \textit{each} frame is given as guidance, while our model only has the segmentation mask for the \textit{first} frame as guidance.

\subsection{Demo Video}
\label{subsec:demo_video}
We also offer a \href{https://www.youtube.com/watch?v=oih0Zk-UW18}{demo video}. This video showcases more video matting results and a hugging face demo for applicability, both on real-world videos.

\end{document}